\documentclass[english]{article}
\usepackage[preprint,nonatbib]{nips_2018_wider_nonotice}

\usepackage[T1]{fontenc}
\usepackage[latin9]{inputenc}
\usepackage{color}
\usepackage{babel}
\usepackage{verbatim}
\usepackage{url}
\usepackage{amsmath}
\usepackage{amssymb}
\usepackage{graphicx}
\usepackage{setspace}
\usepackage{algorithm}
\usepackage{algpseudocode}
\usepackage{hyperref}
\hypersetup{unicode=true,pdfusetitle,
 breaklinks=false}
\usepackage[hyperpageref]{backref}

\graphicspath{{figures}}

\makeatletter
\newcommand{\be}{\begin{eqnarray}}
\newcommand{\ee}{\end{eqnarray}}
\newcommand{\CB}{{\cal B}}

\allowdisplaybreaks \numberwithin{equation}{section}
\makeatother

\title{An Empirical Model of Large-Batch Training}

\author{
  Sam McCandlish\thanks{Work done as an OpenAI Fellow.} \\
  OpenAI \\
  \texttt{sam@openai.com} \\
\And
  Jared Kaplan\\
Johns Hopkins University, OpenAI \\
  \texttt{jaredk@jhu.edu} \\
\And
  Dario Amodei \\
  OpenAI \\
  \texttt{damodei@openai.com} \\
\And \\
  and the \textbf{OpenAI Dota Team}\thanks{The OpenAI Dota Team (Greg Brockman, Brooke Chan, Przemys\l{}aw Debiak, Christy Dennison, David Farhi,  Rafa\l{} J\' ozefowicz, Jakub Pachocki,   Michael Petrov,  Henrique Pond\' e, Jonathan Raiman,  Szymon Sidor,  Jie Tang, Filip Wolski, and Susan Zhang) performed measurements of the reinforcement learning agents they developed for the game Dota 2.  The Dota team's work can be cited as \cite{OpenAI_Five}.}
}

\begin{document}
\maketitle

\begin{abstract}
In an increasing number of domains it has been demonstrated that deep learning models can be trained using relatively large batch sizes without sacrificing data efficiency. However the limits of this massive data parallelism seem to differ from domain to domain, ranging from batches of tens of thousands in ImageNet to batches of millions in RL agents that play the game Dota 2.
To our knowledge there is limited conceptual understanding of why these limits to batch size differ or how we might choose the correct batch size in a new domain.
In this paper, we demonstrate that a simple and easy-to-measure statistic called the \emph{gradient noise scale} predicts the largest useful batch size across many domains and applications, including a number of supervised learning datasets (MNIST, SVHN, CIFAR-10, ImageNet, Billion Word), reinforcement learning domains (Atari and Dota), and even generative model training (autoencoders on SVHN). We find that the noise scale increases as the loss decreases over a training run and depends on the model size primarily through improved model performance. Our empirically-motivated theory also describes the tradeoff between compute-efficiency and time-efficiency, and provides a rough model of the benefits of adaptive batch-size training.
\end{abstract}

\newpage\tableofcontents{}

\section{Introduction}
The last few years have seen a rapid increase in the amount of computation used to train deep learning models \cite{AI_And_Compute}. A major enabler as well as a limiting factor in this growth has been parallelism -- the extent to which a training process can be usefully spread across multiple devices. Regardless of how much total computation is available, if model training cannot be sufficiently parallelized, then it may take too much serial time and therefore may be practically infeasible.

A very common source of parallelism in deep learning has been data parallelism, which involves splitting a batch of data across multiple devices and then aggregating and applying the resulting gradients. Data parallelism requires fast communication between devices, but also requires that large batches are algorithmically effective in accelerating learning. Recently, a number of papers have shown empirically that on specific datasets or tasks, large batch sizes can achieve almost linear speed-ups in training without substantially harming sample efficiency or generalization. For example, batch sizes of 8 thousand \cite{1706.02677}, 16 thousand \cite{1711.00489}, 32 thousand \cite{1708.03888,1709.05011, 1711.04325}, and even 64 thousand \cite{1807.11205} examples have been effectively employed to train ImageNet, and batch sizes of thousands have been effective for language models and generative models \cite{1806.00187,1808.01371,1809.11096}. This phenomenon is not confined to supervised learning: in reinforcement learning, batch sizes of over a million timesteps (with tens of thousands of environments running in parallel) have been used in a Dota-playing agent \cite{OpenAI_Five}, and even in simple Atari environments batch sizes of several thousand timesteps have proved effective \cite{1801.02852,1803.00933,1803.02811}. These discoveries have allowed massive amounts of data and computation to be productively poured into models in a reasonable amount of time, enabling more powerful models in supervised learning, RL, and other domains.

However, for a given dataset and model, there is little to guide us in predicting how large a batch size we can feasibly use, why that number takes a particular value, or how we would expect it to differ if we used a different dataset or model. For example, why can we apparently use a batch size of over a million when training a Dota agent, but only thousands or tens of thousands when training an image recognition model? In practice researchers tend to simply experiment with batch sizes and see what works, but a downside of this is that large batch sizes often require careful tuning to be effective (for example, they may require a warmup period or an unusual learning rate schedule), so the fact that it is possible to use a large batch size can remain undiscovered for a long time. For example, both the Atari and ImageNet tasks were for several years conventionally run with a substantially smaller batch size than is now understood to be possible. Knowing ahead of time what batch size we expect to be effective would be a significant practical advantage in training new models.

\begin{figure}
\noindent \centering{} \includegraphics[width=0.95\textwidth]{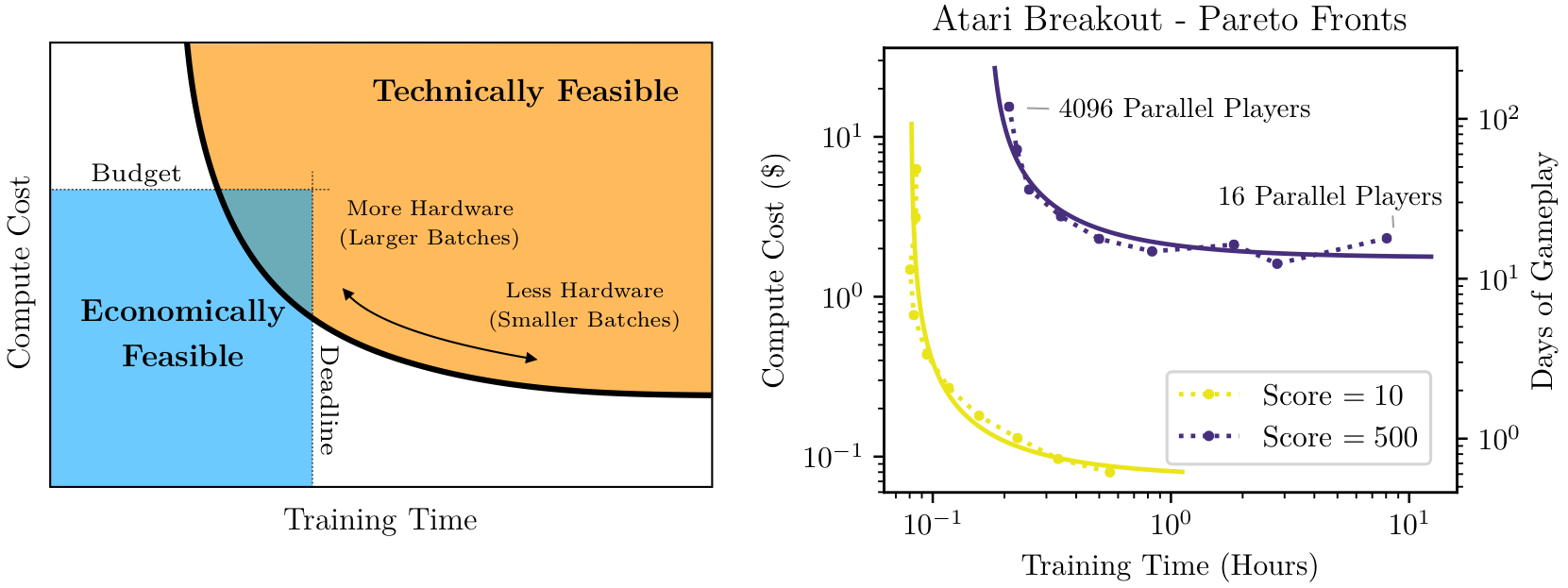}  \caption{The tradeoff between time and compute resources spent to train a model to a given level of performance takes the form of a Pareto frontier (left).  Training time and compute cost are primarily determined by the number of optimization steps and the number of training examples processed, respectively.  We can train a model more quickly at the cost of using more compute resources. On the right we show a concrete example of the Pareto frontiers obtained from training a model to solve the Atari Breakout game to different levels of performance.  The cost and training time depend on the computing architecture and are shown approximately.  \label{fig:tradeoff-curve}}
\end{figure}

In this paper we attempt to answer some of these questions. We measure a simple empirical statistic, the gradient noise scale\footnote{Similar metrics have appeared previously in the literature.  We discuss related work in Section \ref{sec:RelatedWork}.} (essentially a measure of the signal-to-noise ratio of gradient across training examples), and show that it can approximately predict the largest efficient batch size for a wide range of tasks. Our model also predicts a specific shape for the compute/time tradeoff curve, illustrated in Figure \ref{fig:tradeoff-curve}.  Our contributions are a mix of fairly elementary theory and extensive empirical testing of that theory.

On the conceptual side, we derive a framework which predicts, under some basic assumptions, that training should parallelize almost linearly up to a batch size equal to the noise scale, after which there should be a smooth but relatively rapid switch to a regime where further parallelism provides minimal benefits. Additionally, we expect that the noise scale should increase during training as models get more accurate, and should be larger for more complex tasks, but should not have a strong dependence on model size per se. We also provide an analysis of the efficiency gains to be expected from dynamically adjusting the batch size according to noise scale during training. Finally, we predict that, all else equal, the noise scale will tend to be larger in complex RL tasks due to the stochasticity of the environment and the additional variance introduced by the credit assignment problem.

On the empirical side, we verify these predictions across 8 tasks in supervised learning, RL, and generative models, including ImageNet, CIFAR-10, SVHN, MNIST, BillionWord, Atari, 
OpenAI's Dota agent \cite{OpenAI_Five}, and a variational autoencoder for images. For each of these tasks we demonstrate that the noise scale accurately predicts the largest usable batch size (at the order of magnitude level) and that gains to parallelism degrade in the manner predicted by theory. We also show that the noise scale increases over the course of training and demonstrate efficiency gains from dynamic batch size tuning.
The noise scale eventually becomes larger for more performant models, but this appears to be caused by the fact that more performant models simply achieve a better loss. 

The rest of this paper is organized as follows. In Section \ref{sec:ParallelismandNoise}, we derive a simple conceptual picture of the noise scale, data parallelism, and batch sizes, and explain what it predicts about optimal batch sizes and how they vary over the course of training and across tasks. We build on this analysis to study training efficiency in Section \ref{sec:Efficiency}. Then in Section \ref{sec:EmpiricalResults} we empirically test the predictions in Section \ref{sec:ParallelismandNoise} and explore how the noise scale varies with dataset, model size, and learning paradigm (supervised learning vs RL vs generative models). Section \ref{sec:RelatedWork} describes related work and Section \ref{sec:Discussion} discusses the implications of these results and possible future experiments.

\section{Theory and Predictions for the Gradient Noise Scale}

\label{sec:ParallelismandNoise}

\subsection{Intuitive Picture}

\label{subsec:IntuitivePicture}

\begin{figure}
\noindent \centering{}\includegraphics[height=0.25\paperwidth]{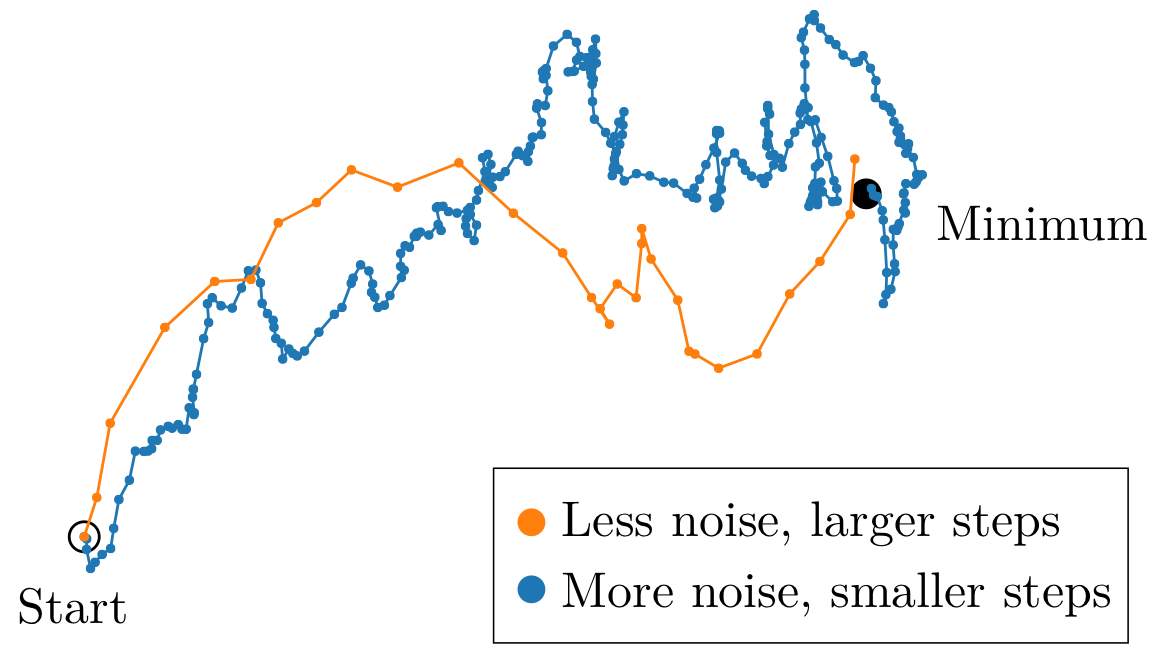}\caption{Less noisy gradient estimates allow SGD-type optimizers to take larger steps, leading to convergence in a smaller number of iterations.  As an illustration, we show two optimization trajectories using momentum in a quadratic loss, with different step sizes and different amounts of artificial noise added to the gradient. \label{fig:noise-illustration}}
\end{figure}
Before working through the details of the gradient noise scale and the batch size, it is useful to present the intuitive picture. Suppose we have a function we wish to optimize via stochastic gradient descent (SGD). There is some underlying true optimization landscape, corresponding to the loss over the entire dataset (or, more abstractly, the loss over the distribution it is drawn from). When we perform an SGD update with a finite batch size, we're approximating the gradient to this true loss. How should we decide what batch size to use?

When the batch size is very small, the approximation will have very high variance, and the resulting gradient update will be mostly noise. Applying a bunch of these SGD updates successively will average out the variance and push us overall in the right direction, but the individual updates to the parameters won't be very helpful, and we could have done almost as well by aggregating these updates in parallel and applying them all at once (in other words, by using a larger batch size).  For an illustrative comparison between large and small batch training, see Figure \ref{fig:noise-illustration}.

By contrast, when the batch size is very large, the batch gradient will almost exactly match the true gradient, and correspondingly two randomly sampled batches will have almost the same gradient.  As a result, doubling the batch size will barely improve the update -- we will use twice as much computation for little gain.

Intuitively, the transition between the first regime (where increasing the batch size leads to almost perfectly linear speedups) and the second regime (where increasing the batch size mostly wastes computation) should occur roughly where the noise and signal of the gradient are balanced -- where the variance of the gradient is at the same scale as the gradient itself\footnote{Note that these considerations are completely agnostic about the size of the dataset itself.}. Formalizing this heuristic observation leads to the noise scale.

The situation is shown pictorially in Figure \ref{fig:tradeoff-curve}. For a given model, we'd like to train it in as little wall time as possible (x-axis) while also using as little total computation as possible (y-axis) -- this is the usual goal of parallelization. Changing the batch size moves us along a tradeoff curve between the two. Initially, we can increase the batch size without much increase in total computation, then there is a ``turning point'' where there is a substantive tradeoff between the two, and finally when the batch size is large we cannot make further gains in training time. In the conceptual and experimental results below, we formalize these concepts and show that the bend in the curve (and thus the approximate largest effective batch size) is in fact set roughly by the noise scale.

\subsection{Gradients, Batches, and the Gradient Noise Scale}

\label{sec:gradients}

We'll now formalize the intuitions described in Section \ref{subsec:IntuitivePicture}. Consider a model, parameterized by variables $\theta\in\mathbb{R}^{D}$, whose performance is assessed by a loss function $L\left(\theta\right)$. The loss function is given by an average over a distribution $\rho\left(x\right)$ over data points $x$. Each data point $x$ has an associated loss function $L_{x}\left(\theta\right)$, and the full loss is given by $L\left(\theta\right)=\mathbb{E}_{x\sim\rho}\left[L_{x}\left(\theta\right)\right]$\footnote{In the context of reinforcement learning, the loss could be the surrogate policy gradient loss, and the distribution $\rho$ would be nonstationary.}.

We would like to minimize $L\left(\theta\right)$ using an SGD-like optimizer, so the relevant quantity is the gradient $G\left(\theta\right)=\nabla L\left(\theta\right)$. However, optimizing $L\left(\theta\right)$ directly would be wasteful if not impossible, since it would require processing the entire data distribution every optimization step. Instead, we obtain an estimate of the gradient by averaging over a collection of samples from $\rho$, called a batch:
\begin{equation}
G_{{\rm est}}\left(\theta\right)=\frac{1}{B}\sum_{i=1}^{B}\nabla_{\theta}L_{x_{i}}\left(\theta\right);\qquad x_{i}\sim\rho
\end{equation}
This approximation forms the basis for stochastic optimization methods such as mini-batch stochastic gradient descent (SGD) and Adam \cite{kingma2014adam}. The gradient is now a random variable whose expected value (averaged over random batches) is given by the true gradient. Its variance scales inversely with the batch size $B$\footnote{This is strictly true only when training examples are sampled independently from the same data distribution. For example, when batches are sampled without replacement from a dataset of size $D$, the variance instead scales like $\left(\frac{1}{B}-\frac{1}{D}\right)$. For simplicity, we restrict ourself to the case where $B\ll D$ or where batches are sampled \emph{with} replacement, but our conclusions can be altered straightforwardly to account for correlated samples.}:
\begin{align}
\mathbb{E}_{x_{1\cdots B}\sim\rho}\left[G_{{\rm est}}\left(\theta\right)\right] & =G\left(\theta\right)\nonumber \\
{\rm cov}_{x_{1\cdots B}\sim\rho}\left(G_{{\rm est}}\left(\theta\right)\right) & =\frac{1}{B}\Sigma\left(\theta\right),\label{eq:gradient-estimate-variance}
\end{align}
where the per-example covariance matrix is defined by
\begin{align}
\Sigma\left(\theta\right) & \equiv{\rm cov}_{x\sim\rho}\left(\nabla_{\theta}L_{x}\left(\theta\right)\right)\nonumber \\
 & =\mathbb{E}_{x\sim\rho}\left[\left(\nabla_{\theta}L_{x}\left(\theta\right)\right)\left(\nabla_{\theta}L_{x}\left(\theta\right)\right)^{T}\right]-G\left(\theta\right)G\left(\theta\right)^{T}.\label{eq:per-sample-gradient-covariance}
\end{align}

The key point here is that the minibatch gradient gives a noisy estimate of the true gradient, and that larger batches give higher quality estimates. We are interested in how useful the gradient is for optimization purposes as a function of $B$, and how that might guide us in choosing a good $B$. We can do this by connecting the noise in the gradient to the maximum improvement in true loss that we can expect from a single gradient update. To start, let $G$ denote the true gradient and $H$ the true Hessian at parameter values $\theta$.  If we perturb the parameters $\theta$ by some vector $V$ to $\theta - \epsilon V$, where $\epsilon$ is the step size, we can expand true loss at this new point to quadratic order in $\epsilon$:%
\begin{figure}
\noindent \centering{}\includegraphics[height=0.28\textwidth]{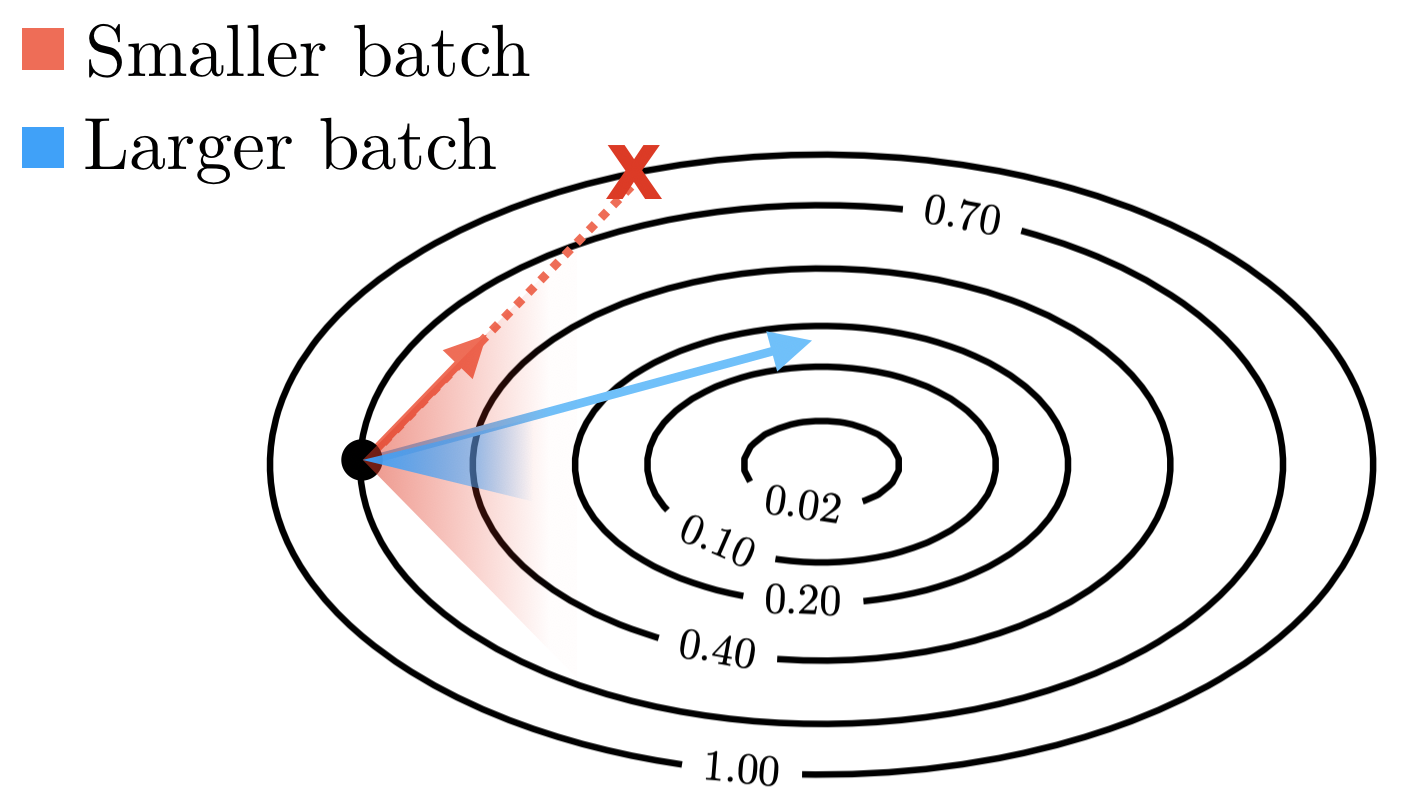}~~~~~\includegraphics[height=0.28\textwidth]{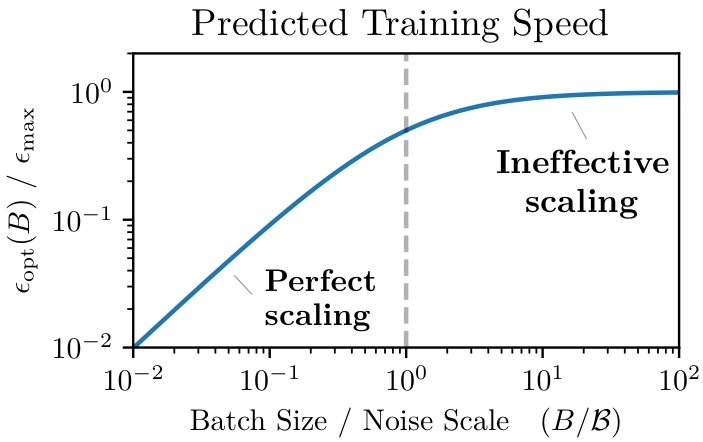}\caption{Larger batch sizes yield estimated gradients that are closer to the true gradient, on average. Larger step sizes can be used when the estimated gradient is closer to the true gradient, so more progress can be made per step. \textbf{Left: }A large step size used with a small batch size can lead to instability, as illustrated for a quadratic loss. \textbf{Right: }Equation \ref{eq:OptimalStepwithNoise} predicts that the `turning point' after which larger batch sizes become less helpful is the noise scale $\mathcal{B}$, where the training speed drops to 50\% of the maximum possible. \label{fig:better-gradient-illustration}}
\end{figure}%
\begin{equation}
L\left(\theta-\epsilon V\right)\approx L\left(\theta\right)-\epsilon G^{T}V+\frac{1}{2}\epsilon^{2}V^{T}HV.\label{eq:quadratic}
\end{equation}
If we had access to the noiseless true gradient $G$ and used it to perturb the parameters, then Equation \ref{eq:quadratic} with $V=G$ would be minimized by setting $\epsilon = \epsilon_{\mathrm{max}}\equiv \frac{|G|^{2}}{G^{T}HG}$. However, in reality we have access only to the noisy estimated gradient $G_{{\rm est}}$ from a batch of size $B$, thus the best we can do is minimize the expectation $\mathbb{E}[L\left(\theta-\epsilon G_{{\rm est}}\right)]$ with respect to $\epsilon$. This expected value can be evaluated using Equation \ref{eq:gradient-estimate-variance}:
\begin{equation}
\mathbb{E}[L\left(\theta-\epsilon G_{{\rm est}}\right)] = L\left(\theta\right)-\epsilon |G|^{2}+\frac{1}{2}\epsilon^{2}\left(G^T H G + \frac{{\rm tr}(H \Sigma)}{B} \right ).
\label{eq:expected-noisy-loss}
\end{equation}

Minimizing this equation with respect to $\epsilon$ leads to:
\begin{equation}
\epsilon_{\mathrm{opt}}\left(B\right)={\rm argmin}_{\epsilon}\mathbb{E}\left[L\left(\theta-\epsilon G_{{\rm est}}\right)\right]=\frac{\epsilon_{{\rm max}}}{1+\mathcal{B}_{{\rm noise}}/{B}}\label{eq:OptimalStepwithNoise}
\end{equation}
as the optimal step size, which produces an optimal improvement in the loss from the noisy gradient:
\begin{align}
\Delta L_{\mathrm{opt}}\left(B\right) & =\frac{\Delta L_{{\rm max}}}{1+\mathcal{B}_{{\rm noise}}/{B}};\qquad\Delta L_{{\rm max}}=\frac{1}{2}\frac{|G|^{4}}{G^{T}HG}.\label{eq:batch-equation}
\end{align}
Above, we have defined the \textit{noise scale }as:
\begin{equation}
\mathcal{B}_{{\rm noise}}=\frac{{\rm tr}\left(H\Sigma\right)}{G^{T}HG},\label{eq:noise-scale}
\end{equation}
Note that our definition of the noise scale is independent of the size of the full training set.  If we use a step size larger than twice $\epsilon_{{\rm opt}}$, the loss may \emph{increase}, leading to divergence, as illustrated in Figure \ref{fig:better-gradient-illustration}.

Despite the many unfounded assumptions in the above derivation, we will find that equations \ref{eq:batch-equation} and \ref{eq:noise-scale} provide a helpful guide to the behavior of large-batch training, even when using other optimizers (including momentum, Adam, and RMSProp).

For a discussion of the dependence of the noise scale on the learning rate, see  Appendix \ref{subsec:temperature} on the `temperature' of training.


\subsubsection*{Implications and Simplifications}

Equation \ref{eq:batch-equation} implies that when the batch size is much smaller than the noise scale, $B\ll\mathcal{B}_{{\rm noise}}$, the second term in the denominator dominates the first, so increasing the batch size $B$ linearly increases the progress in loss. This is the small batch regime, where increases in batch size linearly speed up training. By contrast, when $B\gg\mathcal{B}_{{\rm noise}}$, then the first term dominates, so that increasing $B$ has almost no effect on the progress in loss. This is the large batch regime where increases in batch size do not speed up training and simply waste computation; the switch between the two occurs at $B\approx\mathcal{B}_{{\rm noise}}$ (see Figure \ref{fig:better-gradient-illustration}).

The noise scale in Equation \ref{eq:noise-scale} requires some overhead to compute due to the presence of the Hessian $H$. We can estimate it by measuring $\Delta L_{{\rm opt}}\left(B\right)$ using a series of line searches in the direction of a gradient measured with various batch sizes $B$ and fitting the result to Equation \ref{eq:batch-equation}. This allows us to estimate $\mathcal{B}_{{\rm noise}}$ as well as to empirically test whether Equation \ref{eq:batch-equation} actually fits the data (we discuss these local tests more in Section \ref{sec:EmpiricalResults}).

The situation gets even simpler if we make the (unrealistic) assumption that the optimization is perfectly well-conditioned -- that the Hessian is a multiple of the identity matrix. If that is the case, then Equation \ref{eq:noise-scale} reduces to:
\begin{equation}
\mathcal{B}_{{\rm simple}}=\frac{{\rm tr}(\Sigma)}{|G|^{2}},\label{eq:SimplestNoiseScale}
\end{equation}
which says that the noise scale is equal to the sum of the variances of the individual gradient components, divided by the global norm of the gradient\footnote{One might also use preconditioned gradients, obtained for example by dividing  gradient components by the square root of the Adam optimizer's \cite{kingma2014adam} accumulated variances.  We  experimented with this but found mixed results.} -- essentially a measure of how large the gradient is compared to its variance. It is also a measure of the scale at which the estimated and true gradient become close in $L^{2}$ space (having non-trivial dot product) -- the expected normalized $L^{2}$ distance is given by:
\begin{equation}
\frac{\mathbb{E}\left[\left|G_{{\rm est}}-G\right|^{2}\right]}{|G|^{2}}=\frac{1}{B}\frac{{\rm tr}(\Sigma)}{|G|^{2}}=\frac{\mathcal{B}_{{\rm simple}}}{B},
\end{equation}

In practice, we find that $\mathcal{B}_{{\rm simple}}$ and $\mathcal{B}_{{\rm noise}}$ typically differ only by a small constant multiplicative factor, particularly when we employ common training schemes that improve conditioning. In our empirical work we will sometimes compute $\mathcal{B}_{{\rm noise}}$, but will primarily compute $\mathcal{B}_{{\rm simple}}$ instead, as it requires less computational expense.  In Appendix \ref{subsec:noise-scale-measurement}, we provide an extremely simple method to measure this simplified noise scale with negligible overhead in the context of data-parallel training.

\subsection{Predictions for Data/Time Efficiency Tradeoffs}

\label{sec:Efficiency}

Thus far our analysis has only involved a single point in the loss landscape.  But in Section \ref{sec:EmpiricalResults} we will show that Equation \ref{eq:batch-equation} nevertheless predicts the dependence of training speed on batch size remarkably well, even for full training runs that range over many points in the loss landscape.  By averaging Equation \ref{eq:batch-equation} over multiple optimization steps (see Appendix \ref{app:DynamicBS}), we find a simple relationship between training speed and data efficiency:
\begin{equation}
\frac{S}{S_{{\rm min}}}-1  =  \left(\frac{E}{E_{{\rm min}}}-1\right)^{-1}.\label{eq:tradeoff-eqn}
\end{equation}
Here, $S$ and $S_{{\rm min}}$ represent the actual and minimum possible number of steps taken to reach a specified level of performance, respectively, and $E$ and $E_{{\rm min}}$ represent the actual and minimum possible number of training examples processed to reach that same level of performance.  Since we are training at fixed batch size\footnote{We discuss the benefits of dynamically varying the batch size in Appendix \ref{app:DynamicBS}}, we have $E_{{\rm tot}} =B S_{{\rm tot}}$.  We define the \emph{critical batch size} by an empirical fit to the above equation, as
\begin{equation}
\mathcal{B}_{{\rm crit}} = \frac{E_{{\rm min}}}{S_{{\rm min}}}.
\end{equation}
Our model predicts  $\mathcal{B}_{{\rm crit}} \approx \mathcal{B}_{{\rm noise}}$, where $\mathcal{B}_{{\rm noise}}$ is appropriately averaged over training (see Appendix \ref{app:DynamicBS}).  Note that the noise scale can vary significantly over the course of a training run, so the critical batch size also depends on the level of performance to which we train the model.

The resulting tradeoff curve in serial time vs total compute has a hyperbolic shape represented in Figure \ref{fig:tradeoff-curve}.  The goal of optimization is to reach a given level of performance with minimal $S$ and $E$ -- but as depicted in Figure \ref{fig:tradeoff-curve}, there are tradeoffs involved, as very small $S$ may require very large $E$, and vice versa.   When we choose $B = \mathcal{B}_{{\rm crit}}$, the two sides of Equation \ref{eq:tradeoff-eqn} are both $1$, so that training takes twice as many passes through the training data as an optimally data-efficient (small-batch) run would take, and twice as many optimization steps as an optimally time-efficient (large-batch) run would take.

\subsection{Assumptions and Caveats}

The mathematical argument in the previous sections depends on several assumptions and caveats, and it is useful to list these all in one place, in order to help clarify where and why we might expect the quantities in equations \ref{eq:noise-scale} and \ref{eq:SimplestNoiseScale} to be relevant to training:
\begin{enumerate}
\item \textbf{Short-horizon bias:} The picture in Section \ref{sec:gradients} is a strictly local picture -- it tells us how to best improve the loss on the next gradient step. Greedily choosing the best local improvement is generally not the best way to globally optimize the loss (see e.g. \cite{1803.02021}). For example, greedy optimization might perform poorly in the presence of bad local minima or when the landscape is ill-conditioned.  The critical batch size would then be reduced by the extent to which noise is beneficial.
\item \textbf{Poor conditioning:} In poorly conditioned optimization problems, parameter values often oscillate along the large-curvature directions rather than decreasing in a predictable way (see e.g. \cite{goh2017why} and Appendix \ref{sec:BadDeterministicTraining}).  This means that Equation \ref{eq:batch-equation} will not perfectly reflect the amount of optimization progress made per step.  Nevertheless, we will see that it still accurately predicts the \emph{relative} speed of training at different batch sizes via the resulting tradeoff Equation \ref{eq:tradeoff-eqn}.
\item \textbf{Simplified noise scale:} As noted in Section \ref{sec:gradients}, whenever we use the simplified noise scale (Equation \ref{eq:SimplestNoiseScale}) rather than the exact noise scale (Equation \ref{eq:noise-scale}), this number may be inaccurate to the extent that the Hessian is not well-conditioned. Different components of the gradient can have very different noise scales.
\item \textbf{Learning rate tuning:} The arguments in Section \ref{sec:gradients} assume that we take the optimal step size and maximize the expected improvement in loss, Equation \ref{eq:OptimalStepwithNoise}. In practice learning rates are unlikely to be perfectly tuned, so that the actual improvement in loss (and thus the scaling of training with batch size) may not perfectly reflect Equation \ref{eq:batch-equation}.  However, by trying to choose the best learning rate schedules (or by simply doing a grid search) we can reduce this source of error.  In addition, the noise scale depends strongly on the learning rate via a `temperature' of training, though this source of error is small as long as the learning rate is reasonably close to optimal.  We provide a more detailed discussion of this dependence in Appendix \ref{subsec:temperature}.
 \item \textbf{Quadratic approximation:} The Taylor expansion in Equation \ref{eq:quadratic} is only to second order, so if third order terms are important, in either the distribution of gradient samples or the optimization landscape, then this may introduce deviations from our conceptual model, and in particular deviations from Equation \ref{eq:batch-equation}. Intuitively, since parameter updates are local and often quite small we suspect that the previous two sources of error will be more important than this third one.
\item \textbf{Generalization:} The picture in Section \ref{sec:gradients} says nothing about generalization -- it is strictly about optimizing the training loss as a mathematical function. Some papers have reported a ``generalization gap'' in which large batch sizes lead to good training loss but cause a degradation in test loss, apparently unrelated to overfitting \cite{1609.04836,1705.08741}. The arguments in Section \ref{sec:gradients} don't exclude this possibility, but recent work \cite{1811.03600} has found no evidence of a generalization gap when hyperparameters are properly tuned.
\end{enumerate}
Despite these potential issues in our conceptual model, we'll show in Section \ref{sec:EmpiricalResults} that the noise scale is overall a good empirical predictor of the critical batch size. Furthermore, we will see that most training runs fit Equation \ref{eq:tradeoff-eqn} remarkably well.

\subsection{Expected Patterns in the Noise Scale}

\label{sec:ExpectationsforNS}

In the next section we will measure the noise scale for a number of datasets and confirm its properties. However, it is worth laying out a few patterns we would expect it to exhibit on general grounds:
\begin{itemize}
\item \textbf{Larger for difficult tasks:} We expect $\mathcal{B}$ to be larger for more complex/difficult\footnote{To be clear, we do not expect this to be the primary difference between more and less difficult tasks.  Other difficulty metrics such as the intrinsic dimensionality \cite{1804.08838} appear to be unrelated to the amount of gradient noise, though it would be interesting if there were some connection.} tasks, because individual data points will be less correlated, or only correlated in a more abstract way. This may apply both over the course of training on a given dataset (where we may pick the `low-hanging fruit' first, leaving a long tail of more complicated things to learn) or in moving from easier to harder datasets and environments.  In reinforcement learning, we expect environments with sparse rewards or long time horizons to have larger noise scale. We also expect generative models to have smaller $\CB$ as compared to classifiers training on the same dataset, as generative models may obtain more information from each example.
\item \textbf{Growth over training:} $\mathcal{B}$ will grow when the gradient decreases in magnitude, as long as the noise ${\rm tr}(\Sigma)$ stays roughly constant. Since $|G|$ decreases as we approach the minimum of a smooth loss, we would expect $\mathcal{B}$ to increase during neural network training.
\item \textbf{Weak dependence on model size:} The number of model parameters in the neural network cancels in the noise scale, so we do not expect $\mathcal{B}$ to exhibit a strong dependence on model size (at fixed loss). As discussed above, models that achieve better loss will tend to have a higher noise scale, and larger models often achieve better loss, so in practice we do expect larger models to have higher noise scale, but only through the mechanism of achieving better loss.
\item \textbf{Learning rate tuning:} The noise scale will be artificially inflated if the learning rate is too small, due to the `temperature' dependence described in Appendix \ref{sec:Toy-Models-of}. To get a useful measurement of the noise scale, the learning rate needs to be appropriate to the current point in parameter space.
\end{itemize}
The first and last points can be exhibited analytically in toy models (see Appendix \ref{sec:Toy-Models-of}), but we do not expect theoretical analyses to provide a great deal of insight beyond the intuitions above. Instead, we will focus on confirming these expectations empirically.

\subsection{Summary}

To summarize, our model makes the following predictions about large-batch training:
\begin{itemize}
\item The tradeoff between the speed and efficiency of neural network training is controlled by the batch size and follows the form of Equation \ref{eq:tradeoff-eqn}.
\item The critical batch size $\CB_{{\rm crit}}$ characterizing cost/time tradeoffs can be predicted at the order of magnitude level by measuring the gradient noise scale, most easily in the simplified form $\CB_{{\rm simple}}$ from Equation \ref{eq:SimplestNoiseScale}.
\item The noise scale can vary significantly over the course of a training run, which suggests that the critical batch size also depends on the chosen level of model performance.
\item The noise scale depends on the learning rate via the `temperature' of training, but is consistent between well-tuned training runs (see Appendix \ref{subsec:temperature}).
\end{itemize}

\begin{figure}[t!]
\noindent \centering{} \includegraphics[width=0.7\textwidth]{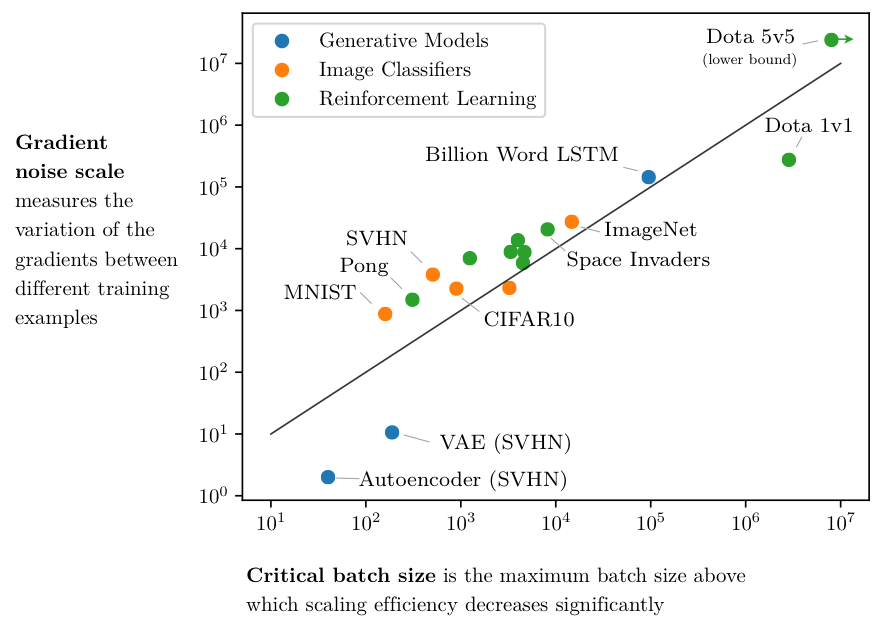} \caption{The ``simple noise scale'' roughly predicts the maximum useful batch size for many ML tasks. We define this ``critical batch size'' to be the point at which compute efficiency drops below 50\% optimal, at which point training speed is also typically 50\% of optimal.  Batch sizes reported in number of images, tokens (for language models), or observations (for games).  We show the critical batch size for a full training run, and the noise scale appropriately averaged over a training run (see Appendix \ref{app:DynamicBS}). Due to resource constraints, for Dota 5v5 we show the batch size used by the OpenAI Dota team as a lower bound for the critical batch size.
\label{fig:noise-scale-summary}}
\end{figure}
\section{Experiments}

\label{sec:EmpiricalResults}

We now test the predictions of Section \ref{sec:ParallelismandNoise} on a range of tasks, including image classification, language modeling, reinforcement learning, and generative modeling.  The tasks range from very simple (MNIST) to very complex (5v5 Dota), which allows us to test our model's predictions in drastically varying circumstances.  Our central experimental test is to compare the prediction made by the gradient noise scale \textbf{$\mathcal{B}_{{\rm simple}}$} for each task, to the actual limits of batch size $\CB_{{\rm crit}}$ found by carefully tuned full training runs at an exhaustive range of batch sizes.  The overall results of this comparison are summarized in Figure \ref{fig:noise-scale-summary}.  We find that the gradient noise scale predicts the critical batch size at the order of magnitude level, even as the latter varies from 20 (for an SVHN autoencoder) to over 10 million (consistent with prior results reported in \cite{OpenAI_Five}).  Details about the hyperparameters, network architectures, and batch size searches are described in Appendix \ref{sec:task-details}. Below we describe the individual tasks, the detailed measurements we perform on each task, and the results of these measurements.

\subsection{Quantities Measured}

\begin{figure}
\noindent \centering{}\includegraphics[height=0.27\textwidth]{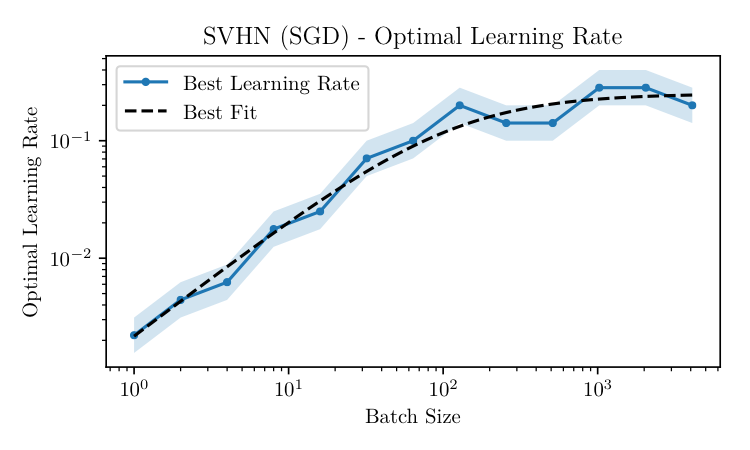}\includegraphics[height=0.27\textwidth]{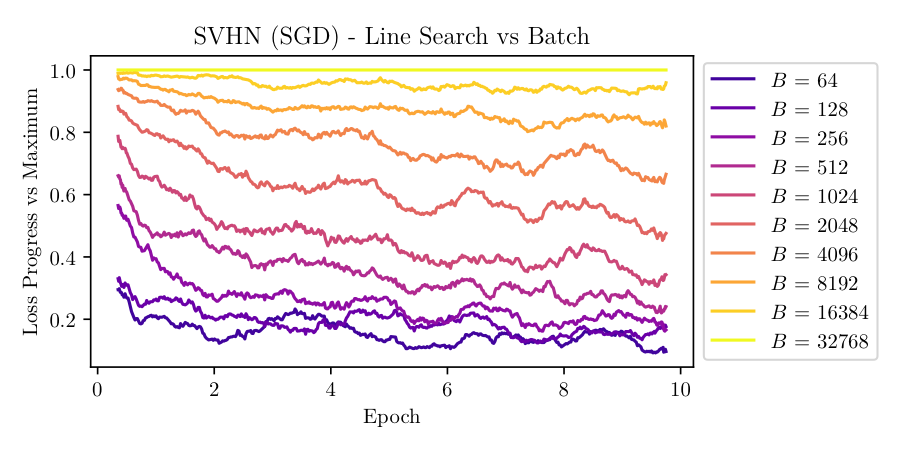} \caption{\textbf{Left: }The optimal learning rate is displayed for a range of batch sizes, for an SVHN classifier trained with SGD. The optimal learning rate initially scales linearly as we increase the batch size, leveling off in the way predicted by Equation \ref{eq:batch-equation}. \textbf{Right:} For a range of batch sizes, we display the average loss progress $\Delta L\left(B\right)$ that can be made from a batch of size $B$ via a line search, normalized by the measured $\Delta L\left(B_{{\rm max}}\right)$. Early in training, smaller batches are sufficient to make optimal progress, while larger batches are required later in training. \label{fig:svhn-optimal-lr}}
\end{figure}

\begin{figure}
\noindent \centering{}\includegraphics[width=1\textwidth]{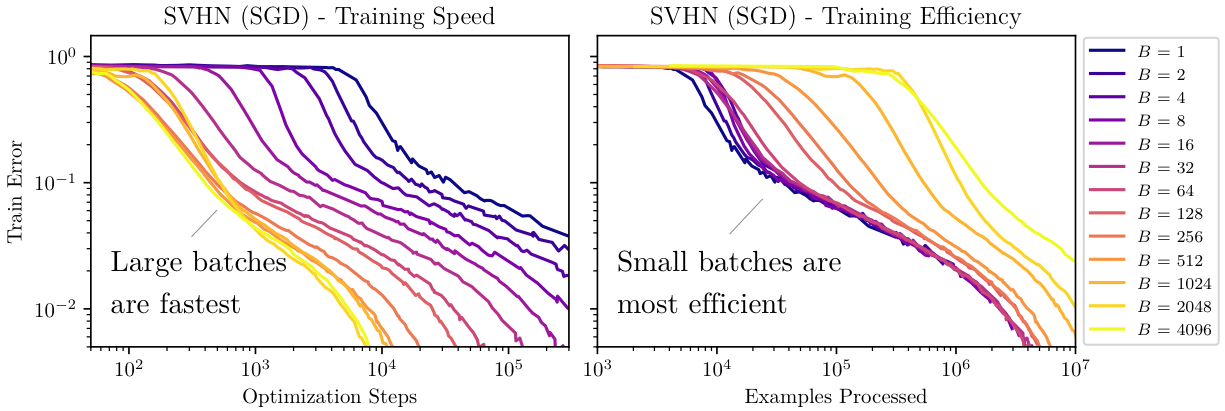} \caption{Training runs for a simple CNN classifier on the SVHN dataset at constant batch sizes. Small batch training is more compute-efficient (right), while large-batch training requires fewer optimizer steps (left). The turning point between time-efficient and compute-efficient training occurs roughly at $B=64$ for the initial phase of training and increases later in training.\label{fig:svhn-epoch-vs-step-comparison}}
\end{figure}
\begin{figure}
\noindent \centering{}\includegraphics[width=0.5\textwidth]{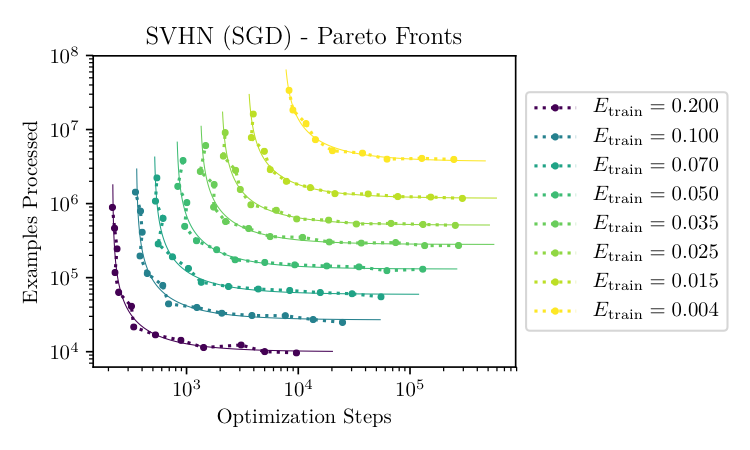}\includegraphics[width=0.5\textwidth]{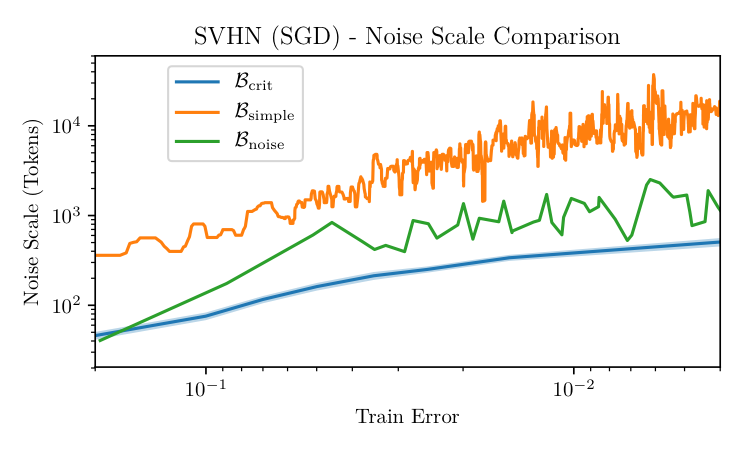}\caption{The tradeoff between time-efficiency and compute-efficiency can be visualized as a Pareto frontier. Each point on the diagram above (left) represents the number of optimizer steps and processed examples needed to achieve a particular level of classification accuracy. Fits to Equation \ref{eq:tradeoff-eqn} are also shown.\label{fig:svhn-pareto-front}}
\end{figure}

In order to test our model we train each task on a range of batch sizes, selecting the optimal constant learning rate separately for each batch size using a simple grid search. Across a range of tasks, we produce the following results and compare with our model:
\begin{itemize}
\item \textbf{Optimal learning rates: }When optimizing with plain SGD or momentum, we find that the optimal learning rate follows the functional form of Equation \ref{eq:OptimalStepwithNoise}, as shown in Figure \ref{fig:svhn-optimal-lr}. For Adam and RMSProp the optimal learning rate initially obeys a power law $\epsilon\left(B\right)\propto B^{\alpha}$ with $\alpha$ between 0.5 and 1.0 depending on the task, then becomes roughly constant. The scale at which the optimal learning rate stops increasing is generally somewhat smaller than the typical noise scale. (See Appendix  \ref{app:ScalingRuleMotivation} for a potential explanation for this power law behavior.)
\item \textbf{Pareto frontiers:} For each batch size, we observe the number of optimization steps and total number of data samples needed to achieve various levels of performance. This allows us to visualize the tradeoff between time-efficiency and compute-efficiency as a Pareto frontier (see Figures \ref{fig:svhn-epoch-vs-step-comparison} and \ref{fig:svhn-pareto-front}). We find that Equation \ref{eq:tradeoff-eqn} fits the shape of these tradeoff curves remarkably well in most cases.
\item \textbf{Critical batch size ($\CB_{{\rm crit}}$)}: We determine the critical batch size over the course of a training run by fitting the Pareto fronts to the functional form of Equation \ref{eq:tradeoff-eqn} (see Figure \ref{fig:svhn-pareto-front}). This quantifies the point at which scaling efficiency begins to drop.  In particular, training runs at batch sizes much less than $\CB_{{\rm crit}}$ behave similarly per training example, while training runs at batch sizes much larger than $\CB_{{\rm crit}}$ behave similarly per optimization step (see Figure \ref{fig:svhn-epoch-vs-step-comparison}).  The critical batch size typically increases by an order of magnitude or more over the course of training.
\item \textbf{Simple noise scale ($\mathcal{B}_{{\rm simple}}$)}: We measure the simple noise scale of Equation \ref{eq:SimplestNoiseScale} over the course of a single training run using the minimal-overhead procedure described in Appendix \ref{subsec:noise-scale-measurement}. Note that some care must be taken to obtain a proper estimate of \textbf{$\mathcal{B}_{{\rm simple}}$} due to its dependence on the learning rate via the `temperature' of training. We find that the noise scale agrees between different well-tuned training runs when compared at equal values of the loss, so  it can be accurately measured at small batch size (see Appendix \ref{sec:Toy-Models-of}).  We also find that, aside from some fluctuations early in training, \textbf{$\mathcal{B}_{{\rm simple}}$} typically predicts the critical batch size at the order of magnitude level (see Figure \ref{fig:svhn-pareto-front}).  The noise scale also typically increases over the course of training, tracking the critical batch size.  To obtain a single value for the noise scale representing a full training run, we average over a training run as described in Appendix \ref{app:DynamicBS}.
\item \textbf{Full noise scale ($\mathcal{B}_{{\rm noise}}$)}: For SVHN trained with SGD, we also measure the full noise scale $\mathcal{B}_{{\rm noise}}$ by performing line searches for gradients obtained by batches of varying size, then fit to the functional form \ref{eq:tradeoff-eqn} (see Figures \ref{fig:svhn-optimal-lr} and \ref{fig:svhn-pareto-front}). This is a somewhat better estimate of $\CB_{{\rm crit}}$ but is less computationally convenient, so we choose to focus on $\mathcal{B}_{{\rm simple}}$ for the remaining tasks.
\end{itemize}


\subsection{Results}

We summarize our findings in Figure \ref{fig:noise-scale-summary}: across many tasks, the typical simple noise scale approximately predicts the batch size at which the returns from increasing scale begin to diminish significantly.   Results for all of the tasks can be found in Appendix \ref{sec:all-results}.  We provide a detailed discussion of our methods in Appendix \ref{app:Methods}.

\subsubsection*{Supervised Learning}

Basic image classification results are pictured in Figure \ref{fig:classification-task-results}.
\begin{itemize}
\item {\bf SVHN }   We train a simple CNN image classifier on the extended SVHN dataset \cite{Netzer2011ReadingDI}.  We display all three of $\CB_{\rm crit}$, $\CB_{\rm simple}$, and $\CB_{\rm noise}$ for SVHN optimized with SGD in Figure \ref{fig:svhn-pareto-front}. We find that $\CB_{\rm noise}$ better-predicts $\CB_{\rm crit}$ as compared to the more naive $\CB_{\rm simple}$.  We  compare to training using  the Adam optimizer \cite{kingma2014adam} in Figure \ref{fig:classification-task-results}, where $\CB_{\rm simple}$ provides a very accurate prediction for $\CB_{\rm crit}$.
\item {\bf MNIST} We train a simple CNN on the MNIST dataset \cite{lecun-mnisthandwrittendigit-2010}  using SGD, and find that $\CB_{\rm simple}$ roughly estimates $\CB_{\rm crit}$, though the latter is significantly smaller.
\item {\bf CIFAR10}  We train a size 32 ResNet  \cite{DBLP:journals/corr/HeZRS15} with Momentum on CIFAR10 \cite{Krizhevsky2009LearningML}  and find that $\CB_{\rm simple}$ predicts $\CB_{\rm crit}$.
\item {\bf ImageNet}  We train a size 50 ResNet \cite{DBLP:journals/corr/HeZRS15}  with Momentum on ImageNet \cite{imagenet_cvpr09}, and use a learning rate schedule that decays the learning rate three times during training.  Due to the schedule, both $\CB_{\rm simple}$, and $\CB_{\rm crit}$ change significantly during training  (see Appendix \ref{subsec:temperature} for a discussion) and must be measured separately at each learning rate.  Results are pictured in Figure \ref{fig:imagenet-results}.  We find that the noise scale varies from 2,000 to 100,000 in the main phase of training, which matches empirical work (e.g. \cite{1807.11205}) showing that constant batch sizes up to 64 thousand can be used without a loss of efficiency.  During the later fine-tuning phase of training, the noise scale increases further to hundreds of thousands and even millions, suggesting that even larger batch sizes may be useful at the very end of training.  Our critical batch sizes are slightly lower (15k vs 64k) than those reported in the literature, but we did not use the latest techniques such as layer-wise adaptive learning rates \cite{1708.03888}.
\end{itemize}

Overall we find that more complex image datasets have larger noise scales in a way that is not directly determined by dataset size.

\subsubsection*{Generative Modeling}

The results for  these tasks are pictured in Figure \ref{fig:generative-modeling-task-results}.
\begin{itemize}
\item {\bf VAE and Autoencoder}   We train a VAE  \cite{1312.6114} and a simple Autoencoder on the SVHN dataset \cite{Netzer2011ReadingDI}; we were motivated to compare these models because VAEs introduce additional stochasticity.  As expected, the VAE had larger $\CB_{\rm crit}$ and $\CB_{\rm simple}$ as compared to the Autoencoder, and both models had much lower $\CB_{\rm simple}$ as compared to SVHN image classifiers.  However, unlike most of the other tasks, for these generative models  $\CB_{\rm simple}$ was significantly \emph{smaller} than $\CB_{\rm crit}$.
\item {\bf Language Modeling}   We train a single-layer LSTM for autoregressive prediction on the Billion Word dataset \cite{1312.3005}, and find good agreement between $\CB_{\rm crit}$ and $\CB_{\rm simple}$.   We also illustrate the dependence on LSTM size in Figure \ref{fig:LSTM-Size-Empirical-BS}, finding that the noise scale is roughly independent of LSTM size at fixed values of the loss, but that larger LSTMs eventually achieve lower values of the loss and a larger noise scale.
\end{itemize}

\subsubsection*{Reinforcement Learning}

\begin{itemize}
\item {\bf Atari}  We train RL agents with the policy gradient algorithm A2C \cite{DBLP:journals/corr/MnihBMGLHSK16} on seven Atari games \cite{DBLP:journals/corr/abs-1207-4708} (Alien, Beamrider,  Breakout, Pong, Qbert, Seaquest, Space Invaders), with results  pictured in Figures \ref{fig:atari-task-results} and \ref{fig:atari-task-results-2}.  The tradeoff curves generally agree well with the prediction of Equation \ref{eq:tradeoff-eqn}, though they are somewhat noisy e.g. for Pong since we do not average over multiple seeds.  For some Atari games, we find some consistent deviation from \ref{fig:atari-task-results-2} at very small batch sizes (see e.g. Beam Rider in Figure \ref{fig:atari-task-results}).  It would be interesting to study this phenomenon further, though this could simply indicate greater sensitivity to other hyperparameters (e.g. momentum) at small batch size.  Overall, we see that patterns in the noise scale match intuition, such as Pong being much easier to learn than other Atari games.
\item {\bf Dota} The OpenAI Dota team has made it possible to train PPO \cite{DBLP:journals/corr/SchulmanWDRK17} agents on both Dota 1v1 and 5v5 environments (the latter being preliminary ongoing work). We vary the two hyperparameters batch size and learning rate on the existing code, experiment setup, and training infrastructure as described in \cite{OpenAI_Five}. The Dota 1v1 environment features two agents fighting in a restricted part of the map (although they are free to walk anywhere) with a fixed set of abilities and skills, whereas Dota 5v5 involves the whole map, 5 heroes on each side, and vastly more configurations in which heroes might engage each other.  This is reflected in the higher noise scale for Dota 5v5 (at least 10 million) relative to Dota 1v1 -- we suspect the higher diversity of situations gives rise to more variance in the gradients.  Due to resource constraints we were not able to measure the Pareto fronts for Dota 5v5, and so we can only report the batch size used by the Dota team and the measured noise scale.
\end{itemize}

Results for the tasks described above were generally within a reasonable margin of state-of-the-art results, though we did not explicitly try to match SOTA or use special algorithmic or architectural tricks.  Our goal was simply to confirm that we were in a reasonably well-performing regime that is typical of ML practice.

For the supervised learning and generative modeling tasks listed above, we have the option of using either training set or test set performance to compare different batch sizes.  For the main results in this paper, we choose train set performance because it is what is directly predicted by our model, and because it is easier to measure in the presence of overfitting.  The choice makes negligible difference in most of our experiments, either because they involve RL or because the datasets are large enough that we don't overfit. On the small datasets MNIST, CIFAR10, and SVHN, overfitting makes measurement of test error more difficult, but we do measure the test error behavior in Appendix \ref{app:generalization}, and both the Pareto fronts and critical batch size generally do not change much.

The fact that the noise scale is consistent between well-tuned training runs suggests that the corresponding optimization trajectories are similar in some sense.  In Appendix \ref{subsec:temperature} we investigate this idea further and relate the noise scale to a characteristic `temperature' of the training process.

\subsubsection*{Model Size Dependence}

\begin{figure}
\noindent \centering{}\includegraphics[width=0.45\textwidth]{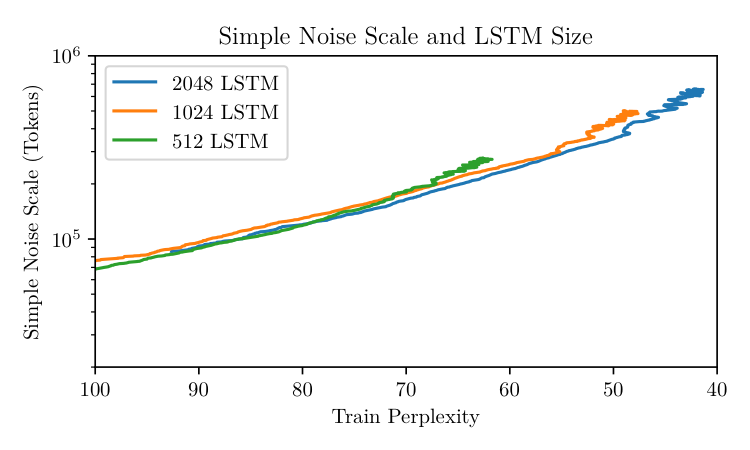}\includegraphics[width=0.45\textwidth]{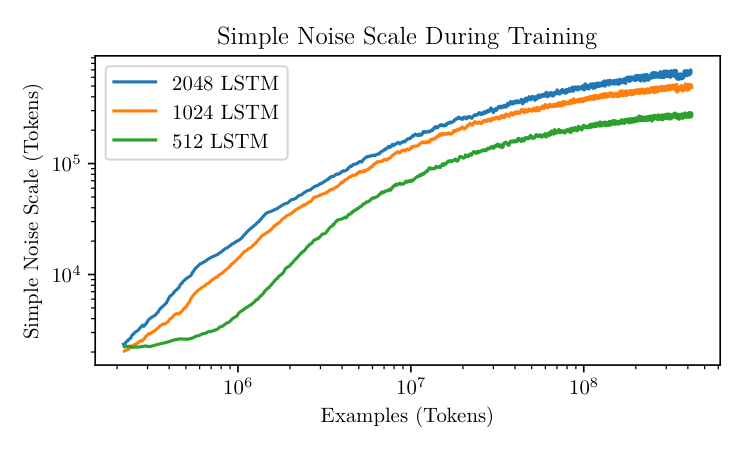} \caption{We show the relationship between training perplexity  and  the simple noise scale (left) for a range of LSTM sizes on the Billion Word dataset. These results show that \emph{at fixed values of the loss, the noise scale does not depend  significantly on model size}.  On the right we show the simple noise scale during training, plotted in terms of the number of tokens processed.  After processing a given number of examples, larger models will tend to have a larger noise scale, but only as a consequence of having  achieved smaller loss.   \label{fig:LSTM-Size-Empirical-BS}}
\end{figure}
The definitions of the noise scales do not have any manifest dependence on the number of parameters in a model. We have conjectured that they will be roughly independent of model size at fixed values of the loss. 

LSTM language models provide a natural test case, since LSTM sizes can be scaled up without qualitatively altering model architecture. As shown in Figure \ref{fig:LSTM-Size-Empirical-BS}, the simple noise scale appears to be roughly independent of model size at a fixed value of the loss.  However, due to their higher performance and lower achieved loss,  larger models eventually reach larger noise scales than smaller models. We do not have specific hypotheses for how the noise scale should vary with model architecture, but interesting results along these lines were recently obtained \cite{1811.03600}.

\section{Related Work}

\label{sec:RelatedWork}
A great deal of prior work has studied large-batch training, investigated versions of the noise scale, explored adaptive batch size and learning rate schedules, and demonstrated that large batch training can be effective on specific datasets.  We attempt to summarize this work below.

Recent papers have probed the limits of large batch training empirically, especially for ImageNet \cite{1706.02677,1709.05011,1807.11205}, in some cases using layer-wise adaptive learning-rates \cite{1708.03888}. More recent work has demonstrated that large batch training can also be applied to RL \cite{1801.02852,OpenAI_Five,1803.02811,1803.00933}. The use of second order optimization methods \cite{ba2016distributed} might increase the utility of data parallelism even further.
A thorough review of large batch training and potential issues with generalization was provided in a very nice recent empirical study \cite{1811.03600} done in parallel with this work. \cite{1811.12941} also systematically studied large-batch training, though it did not tune the learning rate separately for each batch size.

Other recent work has explored the impact of gradient noise on optimization speed and batch size selection.
\cite{conf/icml/SchaulZL13} connected gradient noise and the locally optimal step size to identify an adaptive learning rate.
\cite{1704.04289} derived a sampling distribution for SGD, motivating our definition of `temperature'.
\cite{1710.06451} connected this temperature to the critical batch size, though they predict a dependence on dataset size which we do not observe.
\cite{1703.00810} identified a signal-dominated and noise-dominated phase of training.
\cite{1711.00489} showed that decaying the learning rate and increasing the batch size have the same effect, motivated by the SGD training temperature.
(\cite{1712.02029} also suggested increasing learning rate and batch size together, but with different motivation.)
\cite{1811.02553} empirically investigated the role of gradient noise in reinforcement learning.

The gradient noise scale in particular has also been studied in earlier work to aid in batch size selection.  The noise scale itself is used implicitly in basic statistical techniques for sample size selection (see e.g. \cite{Wiki_noise,NIST_noise}).
\cite{Byrd2012} implicitly uses the gradient noise scale for a theoretical analysis of batch size selection.
\cite{1606.04838,1610.05792,1612.05086} propose adaptive sampling methods based on the gradient noise scale in the context of neural network optimization. 
\cite{1706.05699} analyzed the gradient noise scale for a particular class of functions and related it to the critical batch size, though it predicts a sharp change in learning speed with batch size rather than the smooth change we observe.
\cite{1806.03791} theoretically analyzed the dependence of the gradient noise scale on network width for shallow or linear networks, though they find inconsistent empirical results on neural networks.
\cite{1712.06559} found a formula for the optimization speedup in terms of batch size resembling ours, though their critical batch size depends on smoothness parameters of the loss rather than directly gradient noise.

There has been a variety of work studying the Neural Network loss landscape and using it to draw conclusions about optimal training. Local properties of the loss landscape are not necessarily a good guide to overall optimal training \cite{1803.02021}. The loss tends to be fairly smooth when interpolating between the start and end of training \cite{1412.6544}. But noise may be useful early in training \cite{1511.06807,1706.05699}, perhaps because it leads to minima that generalize better \cite{1609.04836}.

A big-picture motivation for our work was to better understand the scaling of learning with computational and data resources; this question was addressed from the perspective of scaling the model size in \cite{1712.00409}.

Our key contributions include connecting the gradient noise scale to the speed of optimization with a simple model, as well as systematically measuring the critical batch size and noise scale for a variety of tasks.  We also clarify the role of the training temperature in SGD and propose an optimal batch size schedule.

\section{Discussion}

\label{sec:Discussion}

We have shown that the simplified gradient noise scale $\mathcal{B}_{{\rm simple}}$ approximately predicts the actual point of diminishing return on batch size $\CB_{\rm crit}$ on diverse problems where these quantities vary by six orders of magnitude.  Furthermore, the tradeoff curve between total compute and optimization steps associated with changing the batch size has roughly the hyperbolic form predicted by our theory.  Finally, our theory also roughly predicts how the optimal learning rate scales with batch size, although its predictions are not  as precise.

What does the validity of this theory mean, and in what way is it useful?  At the level of a given task, it allows us to use the noise scale from a single run (even an only partially complete run with much smaller batch size, though see caveats about learning rate tuning in the appendix) to estimate the largest useful batch size, and thus reduces the extensive hyperparameter searches that are necessary to find this batch size by trial and error.  It also tells us to expect that larger batch sizes will show diminishing returns in a predictable way that has the same form regardless of the task.

Across tasks, it tells us that the largest useful batch size for a task is likely to be correlated to informal notions of the ``complexity'' of the task, because the noise scale essentially measures how diverse the data is (as seen by the model), which is one aspect of  task complexity.   

We have argued that a specific formula characterizes  the time/compute tradeoff between optimization steps and total data processed in neural network training:
\begin{eqnarray}
\left(\frac{{\rm Optimization\ Steps}}{{\rm Min\ Steps}}-1\right)\left(\frac{{\rm Data\ Examples}}{{\rm Min\ Examples}}-1\right) & = & 1\label{eq:TradeoffDiscussion}
\end{eqnarray}
From this relation we can identify a critical value of the batch size when training to a given value of the loss
\[
\mathcal{B}_{{\rm crit}}({\rm Loss})=\frac{\text{Min Examples}}{\text{Min Steps}}
\]
Training at this critical batch size provides a natural compromise between time and compute, as we take only twice the minimum number of optimization steps and use only twice the minimum amount of data.  The critical batch size represents a turning point, so that for $B > \CB_{\rm crit}$ there are diminishing returns from greater data parallelism.

Our  main goal was to provide a simple way to predict $\CB_{\rm crit}$.  We have shown that it can be estimated as
\begin{eqnarray}
\mathcal{B}_{{\rm crit}}\approx\mathcal{B}_{{\rm simple}}
\end{eqnarray}
where the easily-measured $\mathcal{B}_{{\rm simple}}$ is the ratio of the gradient variance to its squared mean. Theoretical arguments suggest that a more refined quantity, the Hessian-weighted $\CB_{{\rm noise}}$ of Equation \ref{eq:noise-scale}, may provide an even better\footnote{We have also investigated using gradients preconditioned by the Adam optimizer; the results were mixed.} estimate of $\CB_{{\rm crit}}$.

The tradeoff curve of Equation \ref{eq:TradeoffDiscussion} provides a remarkably good fit across datasets, models, and optimizers, and the approximate equality of  $\CB_{{\rm crit}}$ and $\CB_{{\rm simple}}$ holds even as both quantities vary greatly between tasks and training regimes. We have established that as anticipated, both $\CB_{{\rm crit}}$ and $\CB_{{\rm simple}}$ tend to increase significantly during training, that they are larger for more complex tasks, and that they are roughly independent of model size (for LSTMs) at fixed values of the loss. We also saw that image classification has a significantly larger per-image noise scale as compared to generative models training on the same dataset, a fact that could have interesting implications for model-based RL.  In the case of RL, while the noise scale for Dota was roughly a thousand times larger than that of Atari, the total number of optimization steps needed to train a Dota agent is not so much larger \cite{OpenAI_Five}.  Perhaps this suggests that much of the additional compute needed to train more powerful models will be parallelizable.

While $\CB_{{\rm simple}}$ roughly matches $\CB_{{\rm crit}}$ for all datasets, the ratio $\CB_{{\rm simple}}/\CB_{{\rm crit}}$ can vary by about an order of magnitude between tasks. This may not be so surprising, since $\CB_{{\rm simple}}$ does not take into account Hessian conditioning or global aspects of the loss landscape. But it would be very interesting to obtain a better understanding of this ratio. It was smaller than one for the Autoencoder, VAE, and for Dota 1v1, roughly equal to one for LSTMs, and greater than one for both image classification tasks and Atari, and we lack an explanation for these variations.  It would certainly be interesting to study this ratio in other classes of models, and to further explore the behavior of generative models.

Due to its crucial role in data-parallelism, we have focused on the batch size $B$, presuming that the learning rate or effective `temperature' will be optimized after $B$ has been chosen. And our theoretical treatment focused on a specific point in the loss landscape, ignoring issues such as the relationship between early and late training and the necessity of a `warm-up' period. It would be interesting to address these issues, particularly insofar as they may provide  motivation for adaptive batch sizes.

\subsection*{Acknowledgements}
We are grateful to Paul Christiano for initial ideas and discussions about this project. We would like to thank the other members of OpenAI for discussions and help with this project, including Josh Achiam, Danny Hernandez, Geoffrey Irving,  Alec Radford, Alex Ray, John Schulman, Jeff Wu, and Daniel Ziegler.  We would also like to thank Chris Berner, Chris Hesse, and Eric Sigler for their work on our training infrastructure.  We thank Joel Hestness, Heewoo Jun, Jaehoon Lee, and Aleksander Madry for feedback on drafts of this paper.  JK would also like to thank Ethan Dyer for discussions.

\clearpage

\appendix

\section{Methods}
\label{app:Methods}

\subsection{Unbiased Estimate of the Simple Noise Scale with No Overhead}

\label{subsec:noise-scale-measurement}
In this section, we describe a method for measuring the noise scale that comes essentially for free in a data-parallel training environment.

We estimate the noise scale by comparing the norm of the gradient for different batch sizes. From Equation \ref{eq:gradient-estimate-variance}, the expected gradient norm for a batch of size $B$ is given by:
\begin{equation}
\mathbb{E}\left[|G_{{\rm est}}|^{2}\right]=|G|^{2}+\frac{1}{B}{\rm tr}(\Sigma).\label{eq:expected-norm}
\end{equation}
Given estimates of $|G_{{\rm est}}|^{2}$ for both $B=B_{{\rm small}}$ and $B=B_{{\rm big}}$, we can obtain unbiased estimates $|\mathcal{G}|^{2}$ and $\mathcal{S}$ for $|G|^{2}$ and ${\rm tr}(\Sigma)$, respectively:
\begin{align}
|\mathcal{G}|^{2} & \equiv\frac{1}{B_{{\rm big}}-B_{{\rm small}}}\left(B_{{\rm big}}|G_{B_{{\rm big}}}|^{2}-B_{{\rm small}}|G_{B_{{\rm small}}}|^{2}\right)\nonumber\\
\mathcal{S} & \equiv\frac{1}{1/B_{{\rm small}}-1/B_{{\rm big}}}\left(|G_{B_{{\rm small}}}|^{2}-|G_{B_{{\rm big}}}|^{2}\right).\label{eq:easy-noise-measurement}
\end{align}
We can verify with Equation \ref{eq:expected-norm} that $\mathbb{E}\left[|\mathcal{G}|^{2}\right]=|G|^{2}$ and $\mathbb{E}\left[\mathcal{S}\right]={\rm tr}(\Sigma)$.\footnote{Note that when $B_{{\rm small}}=1$ and $B_{{\rm big}}=n$, this becomes the familiar Bessel correction $\frac{n}{n-1}$ to the sample variance.}

Note that the ratio $\mathcal{S}/|\mathcal{G}|^{2}$ is not an unbiased estimator for $\mathcal{B}_{{\rm noise}}$\footnote{In fact $\mathbb{E}\left[x/y\right]\geq\mathbb{E}\left[x\right]/\mathbb{E}\left[y\right]$ in general for positive variables, see e.g. \url{https://en.wikipedia.org/wiki/Ratio_estimator} for details.}. It is possible to correct for this bias, but to minimize complexity we instead ensure that $|\mathcal{G}|^{2}$ has relatively low variance by averaging over many batches.  This is especially important due to the precise cancellation involved in the definition of $|\mathcal{G}|^{2}$.

When training a model using a data parallel method, we can compute $|G_{B_{{\rm small}}}|^{2}$ and $|G_{B_{{\rm big}}}|^{2}$ with minimal effort by computing the norm of gradient before and after averaging between devices. In that case $B_{{\rm small}}$ is the ``local'' batch size before averaging, and $B_{{\rm big}}$ is the ``global'' batch size after averaging.

In practice, to account for the noisiness of $|\mathcal{G}|^{2}$ when computed this way, we calculate $|\mathcal{G}|^{2}$ and $\mathcal{S}$ on every training step and use their values to compute separate exponentially-weighted moving averages.  We tune the exponential decay parameters so that the estimates are stable.  Then, the ratio of the moving averages provides a good estimate of the noise scale.

In our experiments we measure and report the noise scale during training for a single run with a well-optimized learning rate.   Note that while the noise scale measurement is consistent between runs at different batch sizes, it is not consistent at different learning rates (see Appendix \ref{subsec:temperature}).  So, it is important to use a run with a well-tuned learning rate in order to get a meaningful noise scale measurement.

\subsection{Systematic Searches Over Batch Sizes}
\label{sec:learning-rate-scaling}

When doing systematic measurements of how performance scales with batch size (Pareto fronts), we  separately tune the learning rate at each batch size, in order to approximate the ideal batch scaling curve as closely as possible.  We tune the learning rate via the following procedure. For each task, we performed a coarse grid search over both batch size and learning rate to determine reasonable bounds for a fine-grained search. The central value typically followed the form
\begin{equation}
\epsilon_{{\rm central}}\left(B\right)=\frac{\epsilon_{\ast}}{\left(1+B_{\ast}/B\right)^{\alpha}},\label{eq:general-lr-scaling}
\end{equation}
where $\alpha=1$ for SGD or momentum, and $0.5<\alpha<1$ for Adam \cite{kingma2014adam} or RMSProp. Then, we performed an independent grid search for each batch size centered at $\epsilon_{{\rm central}}$, expanding the bounds of the search if the best value was on the edge of the range.

We explain the motivation for Equation \ref{eq:general-lr-scaling} in Appendix \ref{app:ScalingRuleMotivation}.  But regardless of the theoretical motivations, we have found that this scaling rule provides a reasonable starting point for grid searches, though we are not suggesting that they produce precisely optimized learning rates.


\subsection{Pareto Front Measurements}

To produce the Pareto front plots, and thus to measure the important parameter $\CB_{\rm crit}$ for a given dataset and optimizer, we begin by performing a grid search over batch sizes and learning rates, as described in Appendix \ref{sec:learning-rate-scaling}.  With that data in hand, we fix a list of goal values -- either loss, perplexity, or game-score.  For example for SVHN in Figure \ref{fig:svhn-pareto-front} we chose the training classification error values $[0.2, 0.1, 0.07, 0.05, 0.035, 0.025, 0.015, 0.004]$ as the goals.  These were generally chosen to provide a variety of evenly spaced Pareto fronts indicative of  optimization progress.

Then for each  value of the goal, and for each value of the batch size, we identified the number of optimization steps and examples processed for the run (among those in the grid search) that achieved that goal most quickly.  These optimal runs are the data points on the Pareto front plots.  Note that at fixed batch size, different values of the learning rate might be optimal for different values of the goal (this was certainly the case for LSTMs on Billion Word, for example).  Next, for each value of the goal, we used the optimal runs at each value of the batch size to  fit Equation \ref{eq:tradeoff-eqn} to the relation between examples processed and optimization steps.  Note that we performed the fits and extracted the errors in log-space. This was how we produced the lines on the Pareto front plots.

Finally, given this fit, we directly measured $\CB_{\rm crit} = \frac{E_{\rm min}}{S_{\rm min}}$ for each value of the goal, as well as the standard error in this quantity.  This was how we produced the `Noise Scale Comparison' plots, where we compared $\CB_{\rm crit}$ to $\CB_{\rm simple}$.  Errors in $\CB_{\rm crit}$ are standard errors from the fit to Equation \ref{eq:tradeoff-eqn}.  When we report an overall number for $\CB_{\rm crit}$ for a given dataset and optimizer, we are averaging over optimization steps throughout training.

Note that it can be difficult to determine at what point in a training run the model's performance reaches the specified target.  For example, the loss may oscillate significantly, entering and exiting the target region multiple times.  To remedy this issue, we smooth the loss using an exponentially-weighted moving average before checking whether it has reached the target.  The decay parameter of this moving average can affect results noticeably. Though we choose this parameter by hand based on the noisiness of the model's performance, this could be automated using an adaptive smoothing algorithm.

\subsection{Details of Learning Tasks \label{sec:task-details}}

We train a variety of architectures on a variety of ML tasks described below. We use either basic stochastic gradient descent (SGD), SGD with momentum \cite{pmlr-v28-sutskever13}, or the Adam optimizer \cite{kingma2014adam} unless otherwise specified.  We measure and report the noise scale $\CB_{\rm simple}$ during training for a single run of each task with a well-optimized learning rate.

\subsubsection{Classification}

For image classification, we use the following datasets:
\begin{itemize}
\item MNIST handwritten digits \cite{lecun-mnisthandwrittendigit-2010}
\item Street View House Numbers (SVHN) \cite{Netzer2011ReadingDI}
\item CIFAR10 \cite{Krizhevsky2009LearningML}
\item ImageNet \cite{imagenet_cvpr09}
\end{itemize}
For CIFAR10 and ImageNet classification, we use Residual Networks \cite{DBLP:journals/corr/HeZRS15} of size 32 and 50 respectively, based on the TensorFlow Models implementation \cite{TF_Official_Resnet}.  All hyperparameters are unchanged aside from the learning rate schedule;  Instead of decaying the learning rate by a factor of 10 at specified epochs, we decay by a factor of 10 when the training classification error (appropriately smoothed) reaches 0.487, 0.312, and 0.229. For MNIST and SVHN, we use a simple deep network with two sets of convolutional and pooling layers (32 and 64 filters, respectively, with 5x5 filters), one fully-connected hidden layer with 1024 units, and a final dropout layer with dropout rate of 0.4.

We train MNIST models using SGD, SVHN with both SGD and Adam \cite{kingma2014adam}  (with the default parameter settings momentum $=0.9$, $\beta_2 = 0.999$), and CIFAR10 and ImageNet with momentum \cite{pmlr-v28-sutskever13} (with momentum $= 0.9$).

\subsubsection{Reinforcement Learning}

For reinforcement learning, we use the following tasks via OpenAI Gym \cite{1606.01540}:
\begin{itemize}
\item Atari Arcade Learning Environment \cite{DBLP:journals/corr/abs-1207-4708}
\item Dota 1v1 and 5v5 \cite{OpenAI_Five}
\end{itemize}
For Atari
, we use A2C \cite{DBLP:journals/corr/MnihBMGLHSK16} with a pure convolutional 
policy, adapted from OpenAI Baselines \cite{baselines}.  We train using RMSProp with $\alpha = 0.99$ and $\epsilon = 10^{-5}$. We roll out the environments 5 steps at a time, and vary the batch size by varying the number of environments running parallel. At the beginning of training, we randomly step each parallel environment by a random number of steps up to 500, as suggested in \cite{1803.02811}.

As described in \cite{OpenAI_Five} for Dota an asynchronous version of PPO \cite{DBLP:journals/corr/SchulmanWDRK17} was used. The TrueSkill metric \cite{NIPS2006_3079} was used to measure the skill of an agent. Given the fact that the OpenAI Five effort is ongoing, the values for TrueSkill reported in this paper are incomparable with those in \cite{OpenAI_Five}; on this paper's scale, TrueSkill 50 is roughly the level of the best semi-pro players.

\subsubsection{Generative and Language Modeling}

For language modeling, we train a size-2048 LSTM \cite{Hochreiter:1997:LSM:1246443.1246450} on the One Billion Word Benchmark corpus \cite{1312.3005}, using byte pair encoding (BPE) \cite{DBLP:journals/corr/SennrichHB15} with a vocabulary of size 40,000 and a $512$-dimensional embedding space. The LSTMs were trained with Adam using momentum $0.5$, without dropout, with the gradients clipped to norm 10, and with $20$-token sequences. For both training and evaluation LSTM cell states were reset to zero between samples, and so we have reported perplexity for the last token of the 20-token sequences.   We chose to report the batch size in tokens (rather than sequences) because we have found that when the number of sequences and the sequence lengths are varied, both $\CB_{\rm simple}$ and $\CB_{\rm crit}$ depend predominantly on the total number of tokens.

We also trained 1024 and 512 size LSTMs for model size comparison; for the last we used a smaller $256$-dimensional embedding space.  The model size comparison training runs were conducted with a batch size of $1024$ and Adam learning rate of $0.0007$.  The learning rates were chosen from a grid search, which showed that the optimal learning rate did not have a significant dependence on model size.

For generative image modeling, we train a Variational Autoencoder \cite{1312.6114} using the InfoGAN architecture \cite{1606.03657} (see their appendix C.2) on the SVHN dataset. Since VAEs introduce additional stochasticity beyond gradient noise, we also provide training data on a simple autoencoder with the same architecture.

\section{Results for All Tasks}

\begin{table}
\centering{}{
\global\long\def\arraystretch{1.2}%
\begin{tabular}{|ll|rr|rr|}
\cline{3-6}
\multicolumn{1}{l}{} & \multicolumn{1}{l|}{} & \multicolumn{2}{c|}{\textbf{Critical Batch Size}} & \multicolumn{2}{c|}{\textbf{Simple Noise Scale}}\tabularnewline
\multicolumn{1}{l}{} & \multicolumn{1}{l|}{} & Start  & Average  & Start  & Average \tabularnewline
\hline
\multicolumn{2}{|l}{\textbf{Image Classification:}} &  &  &  & \tabularnewline
\hline
 & MNIST                            & 20            & 200           & 50         & 900\tabularnewline
 & SVHN                             & 50            & 500           & 300        & 4,000\tabularnewline
 & CIFAR10                          & 300          & 900           & 400        & 2,000\tabularnewline
 & ImageNet                         & 1,000        & 15,000        & 4,000      & 30,000\tabularnewline
\hline
\multicolumn{2}{|l}{\textbf{Generative and Language Modeling:}} &  &  &  & \tabularnewline
\hline
 & Autoencoder (SVHN)               & 10           & 40            & 2          & 2 \tabularnewline
 & Variational Autoencoder (SVHN)   & 10           & 200           & 10         & 10\tabularnewline
 & Billion Word (per token)         & 700          & 100,000       & 1000       & 150,000\tabularnewline
\hline
\multicolumn{2}{|l}{\textbf{Reinforcement Learning:}} &  &  &  & \tabularnewline
\hline
 & Atari (per frame)               & 100 - 1,000    & 400 - 8,000    & 100-1,000  & 1,000-20,000\tabularnewline
 & Dota 1v1 (per frame)            & 50,000        & 3,000,000     & 100,000    & 300,000 \tabularnewline
 & Dota 5v5 (per frame)            & (not measured)    & >8,000,000 (est.)    & 100,000   & 24,000,000 \tabularnewline
\hline
\end{tabular}}
\bigskip
\caption{We report the simple noise scale, both early in training and averaged over a training run, as well as the critical batch size, both early in the run and at the end of the run.  The noise scale provides a good estimate for the critical batch size throughout training. Batch sizes reported in number of images, tokens (for language models), or observations (for games).  These data are summarized in Figure \ref{fig:noise-scale-summary}. \label{tab:noise-crit-comparison}
}
\end{table}

\label{sec:all-results}

\begin{figure}
\includegraphics[width=0.5\textwidth]{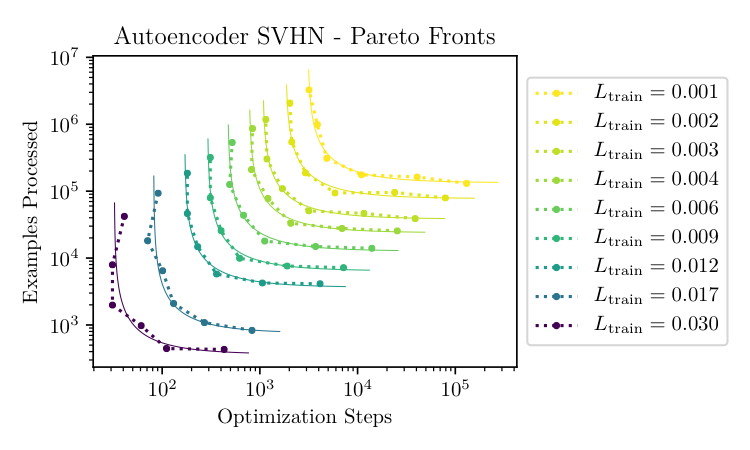}\includegraphics[width=0.5\textwidth]{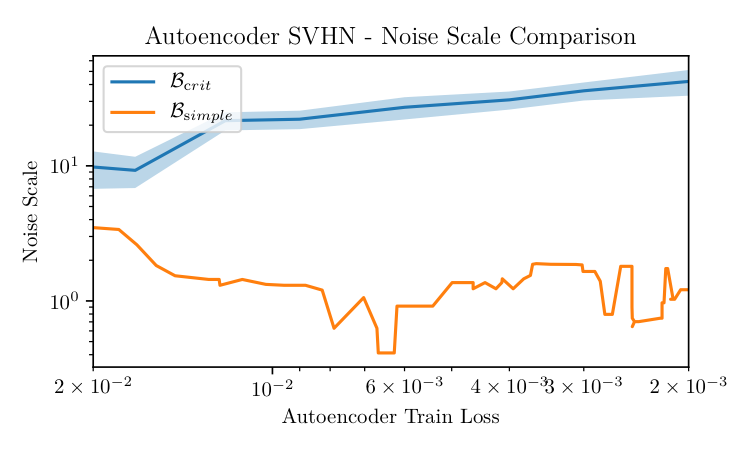}\\
 \includegraphics[width=0.5\textwidth]{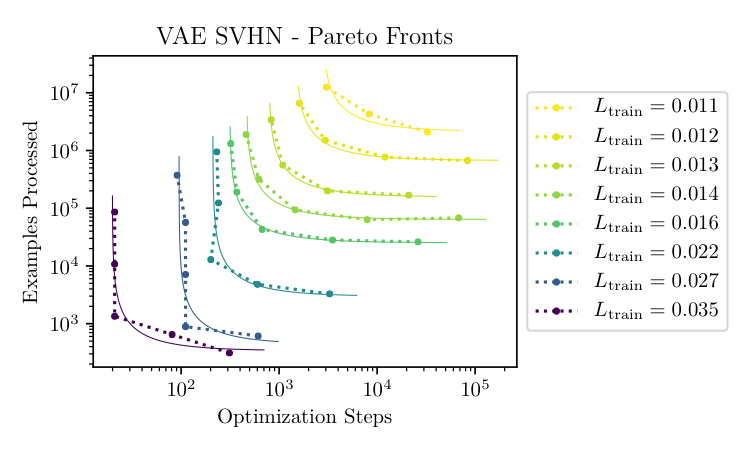}\includegraphics[width=0.5\textwidth]{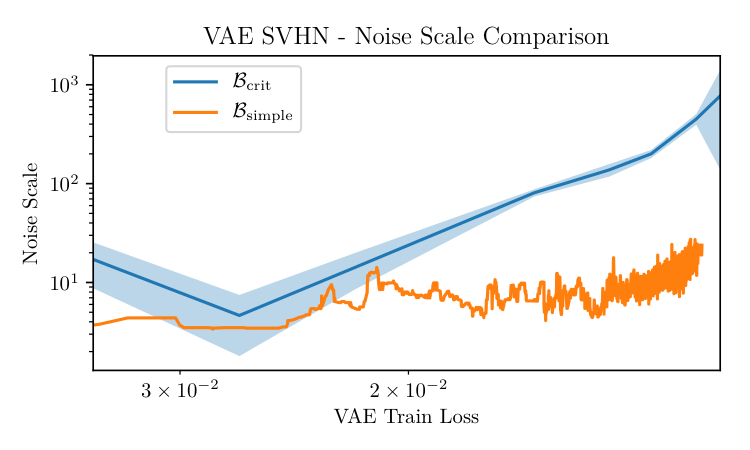}\\
 \includegraphics[width=0.5\textwidth]{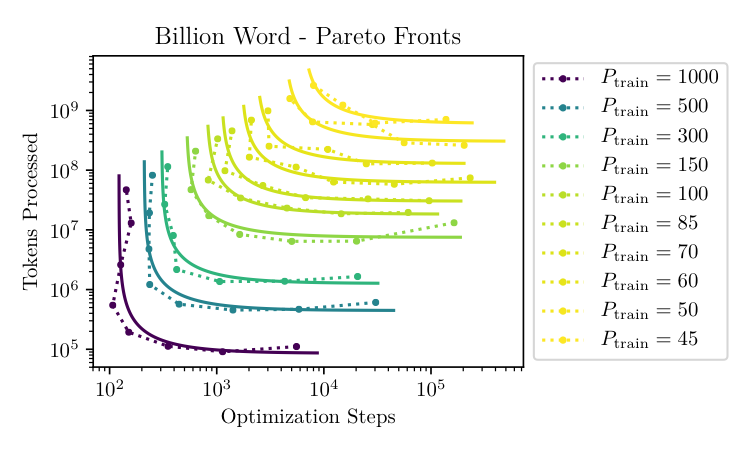}\includegraphics[width=0.5\textwidth]{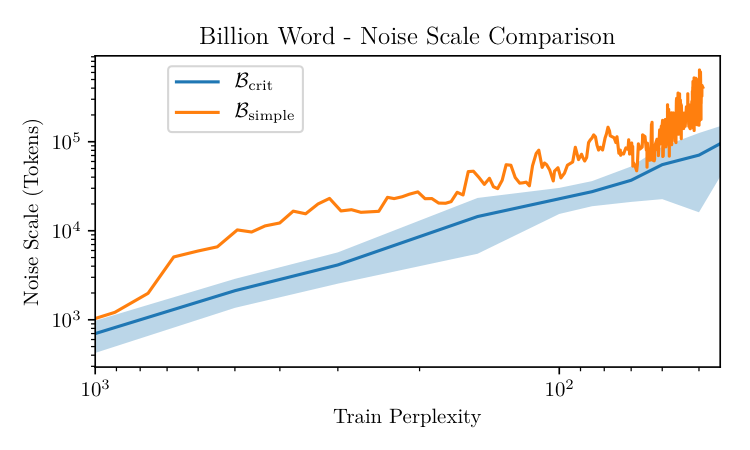}\caption{Scaling behavior of generative and language modeling tasks. \label{fig:generative-modeling-task-results}}
\end{figure}

\begin{figure}
\includegraphics[width=0.5\textwidth]{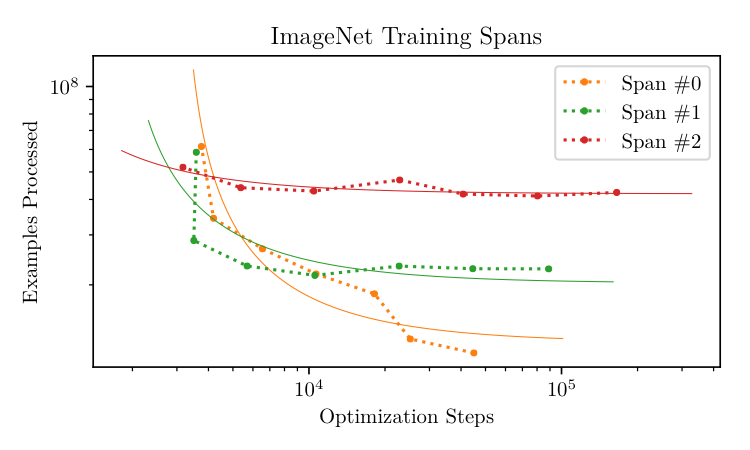}\includegraphics[width=0.5\textwidth]{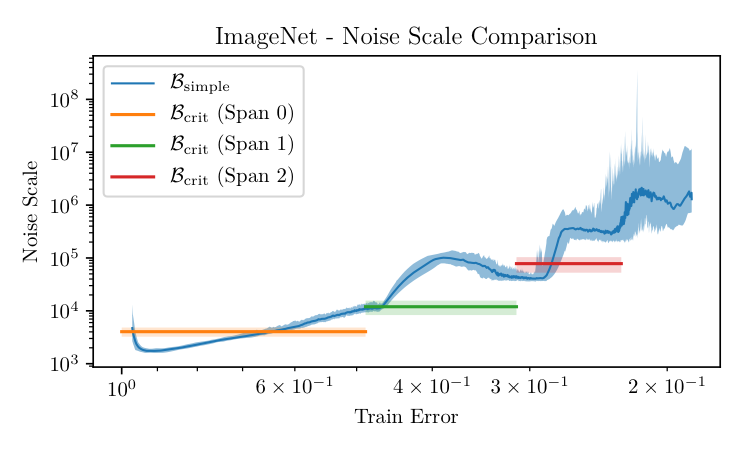}\\
 \caption{For ImageNet, the typical training schedule decays the learning rate by a factor of 10 at 30, 60, and 80 epochs \cite{DBLP:journals/corr/HeZRS15,1706.02677}. To provide a fair comparison between batch sizes, we instead decay by a factor of 10 when the training classification error reaches 0.487, 0.312, and 0.229.  We display Pareto fronts and compute the critical batch size separately for each span.
\label{fig:imagenet-results}}
\end{figure}

\begin{figure}
\includegraphics[width=0.5\textwidth]{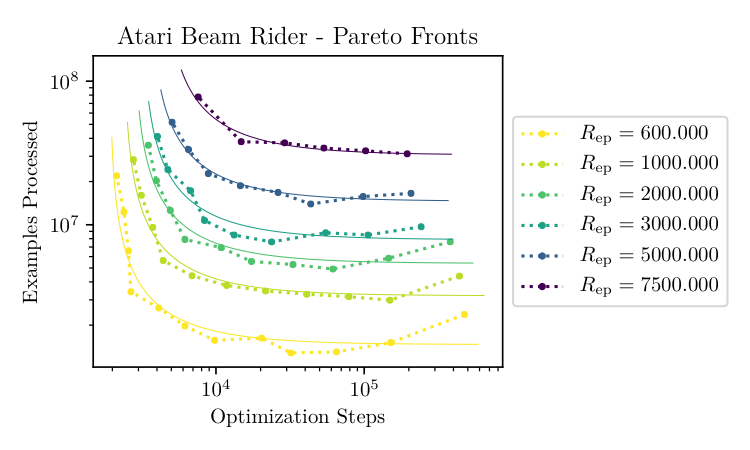}\includegraphics[width=0.5\textwidth]{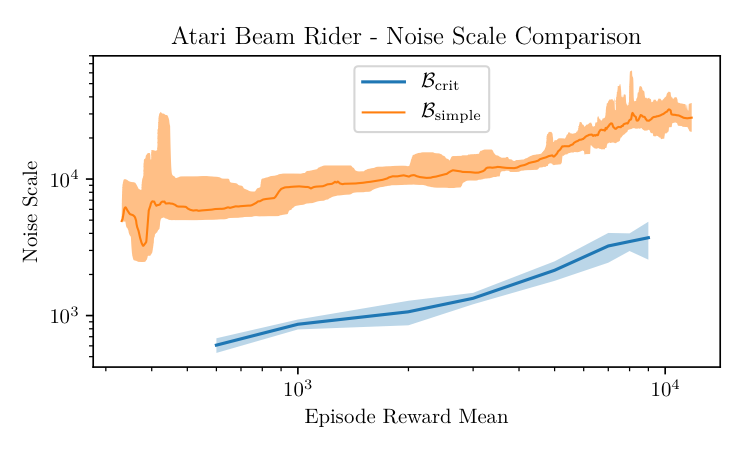}\\
 \includegraphics[width=0.5\textwidth]{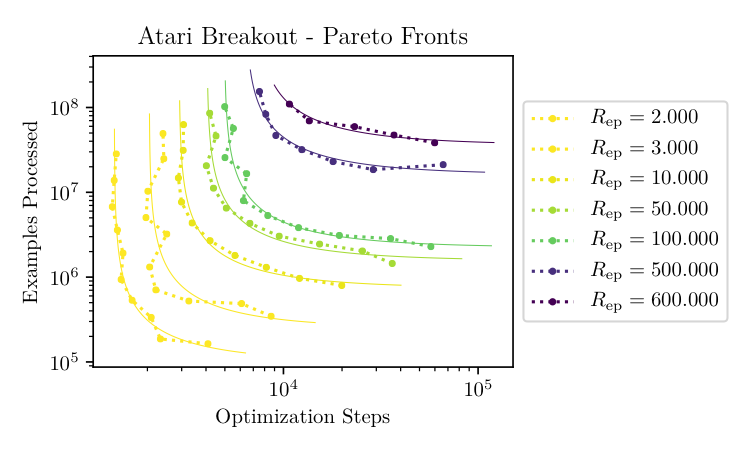}\includegraphics[width=0.5\textwidth]{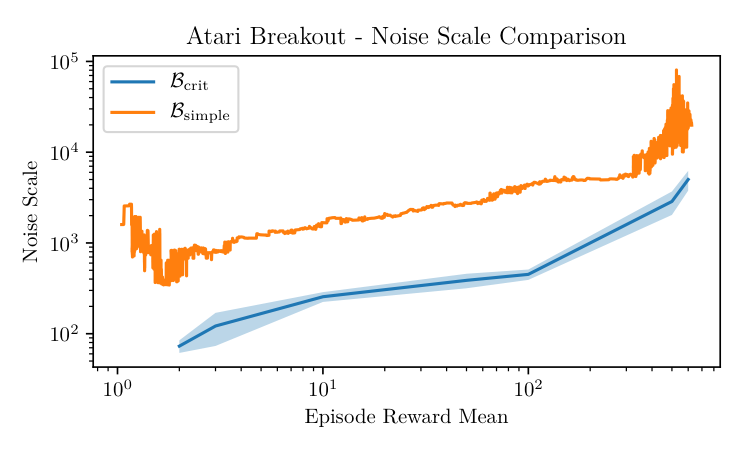}\\
 \includegraphics[width=0.5\textwidth]{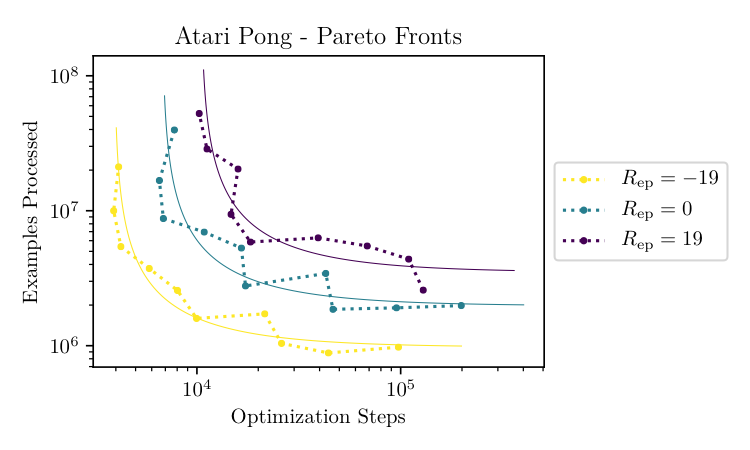}\includegraphics[width=0.5\textwidth]{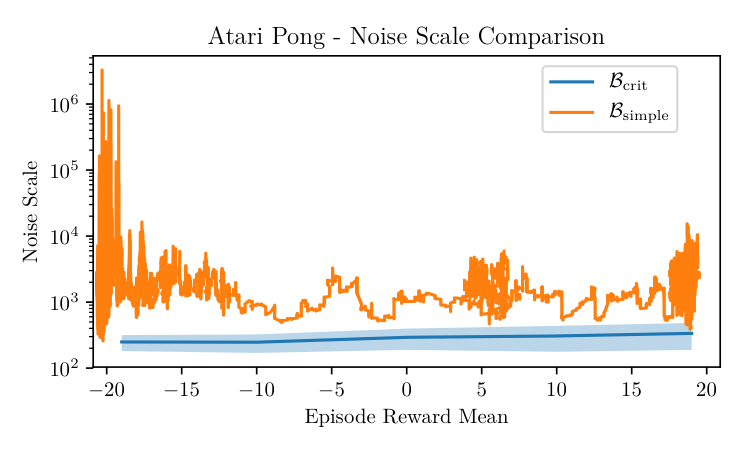}\\
 \includegraphics[width=0.5\textwidth]{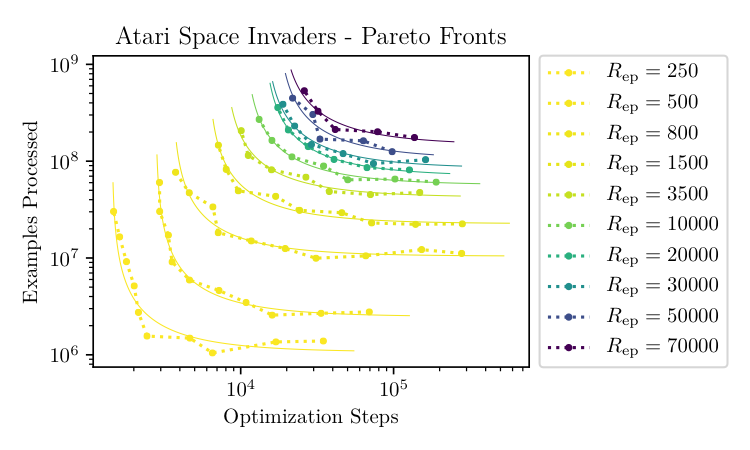}\includegraphics[width=0.5\textwidth]{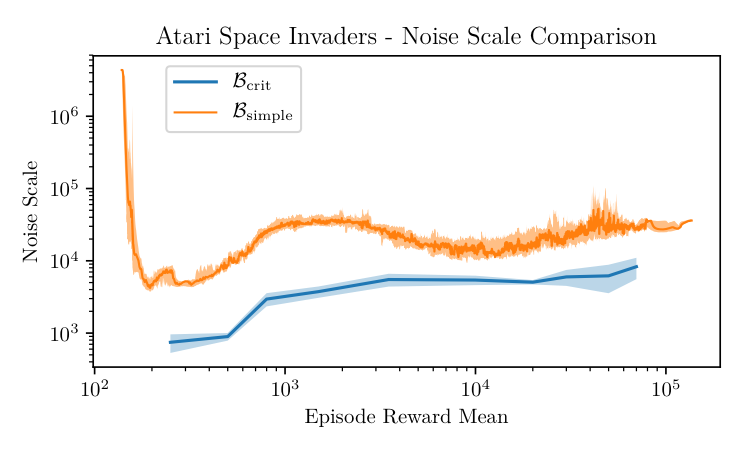} \caption{Scaling behavior of Atari tasks (Beam Rider, Breakout, Pong, and Space Invaders) trained with A2C. \label{fig:atari-task-results}}
\end{figure}

\begin{figure}
\includegraphics[width=0.5\textwidth]{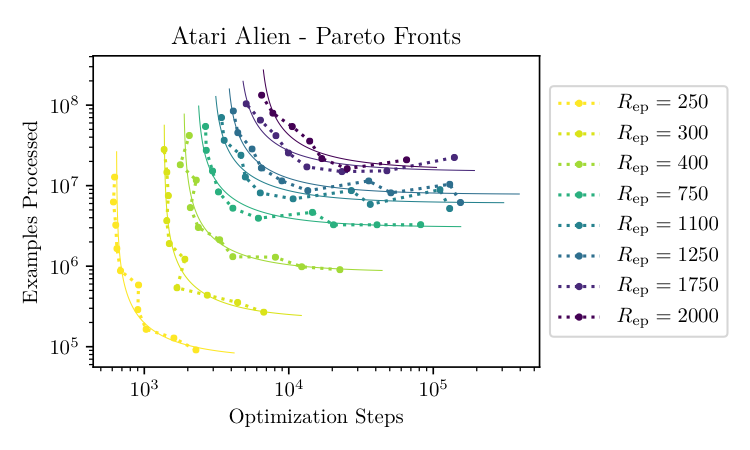}\includegraphics[width=0.5\textwidth]{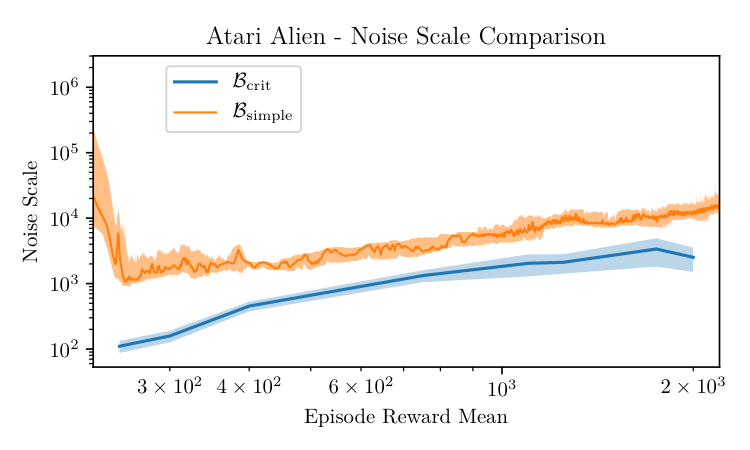}\\
 \includegraphics[width=0.5\textwidth]{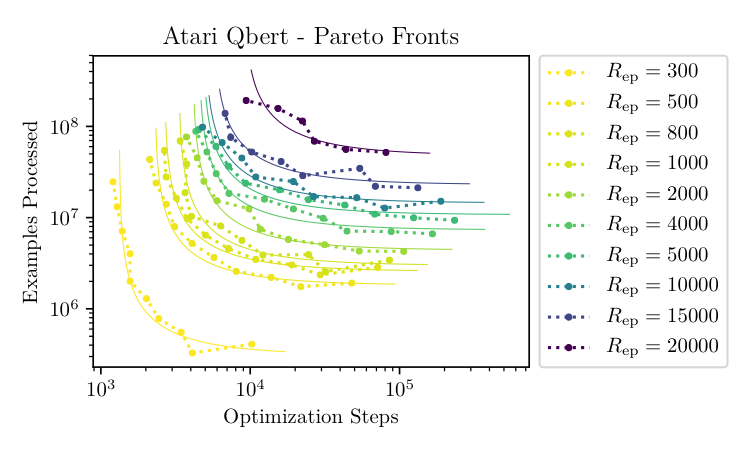}\includegraphics[width=0.5\textwidth]{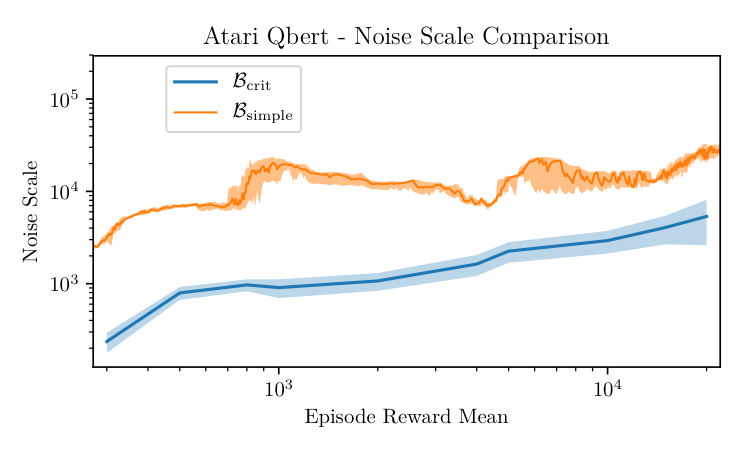}\\
 \includegraphics[width=0.5\textwidth]{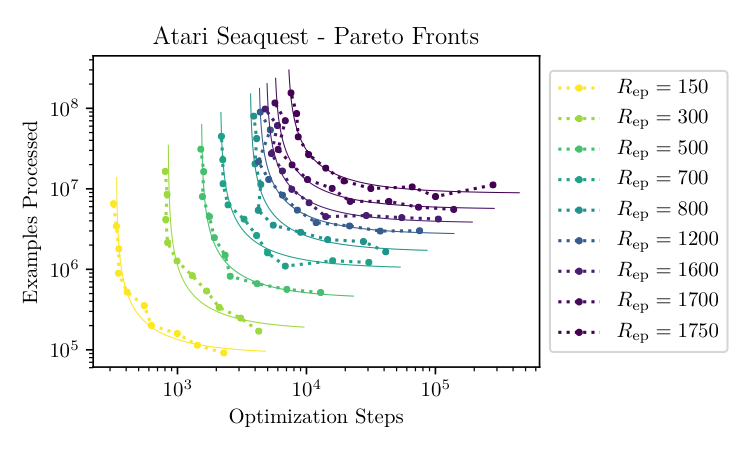}\includegraphics[width=0.5\textwidth]{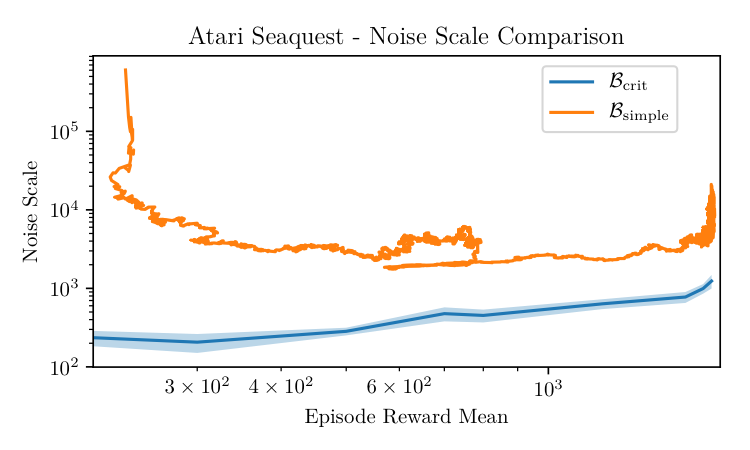}\\
 \caption{Scaling behavior of more Atari tasks (Alien, Qbert, and Seaquest) trained with A2C. \label{fig:atari-task-results-2}}
\end{figure}

\begin{figure}
\includegraphics[width=0.5\textwidth]{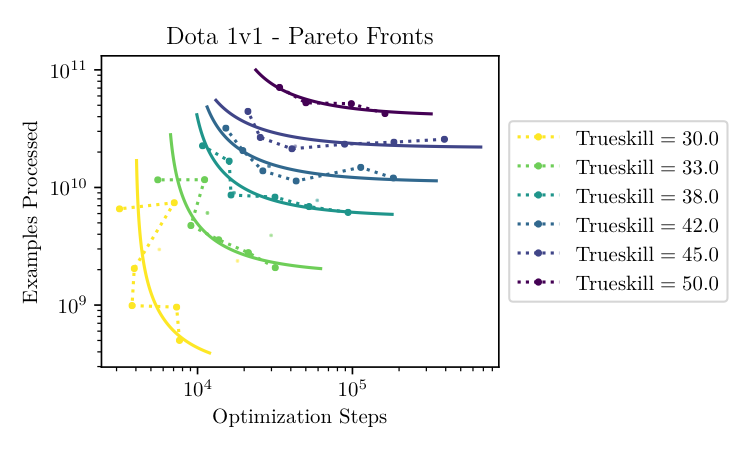}\includegraphics[width=0.5\textwidth]{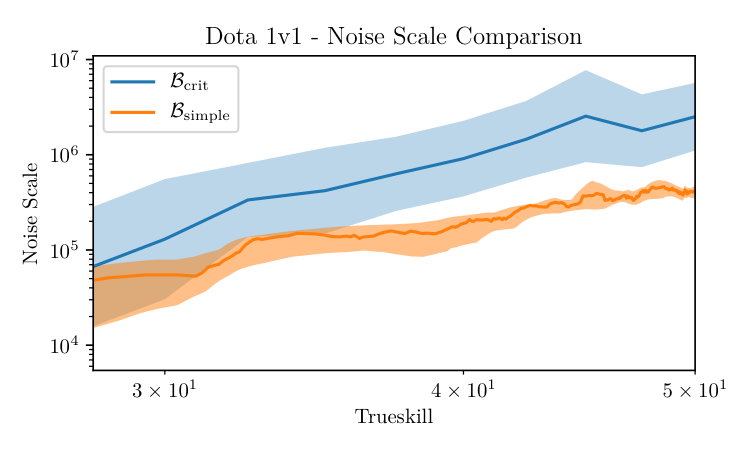} \caption{Scaling behavior for Dota 1v1 \cite{OpenAI_1v1} trained to top-level pro performance.  \label{fig:appdx-dota-results}}
\end{figure}

\begin{figure}
\includegraphics[width=0.5\textwidth]{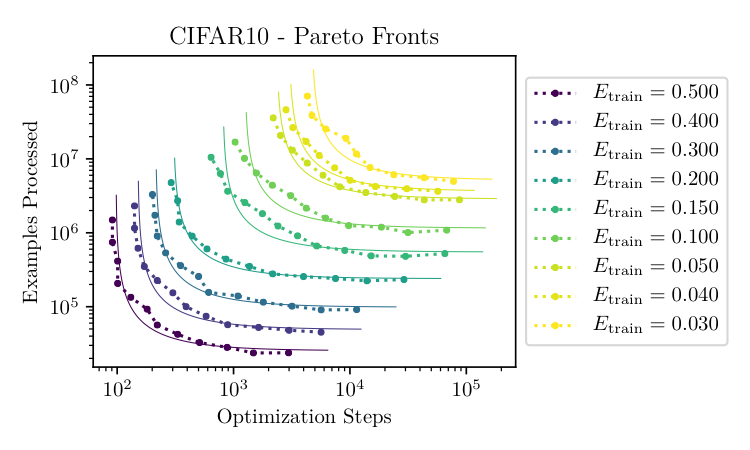}\includegraphics[width=0.5\textwidth]{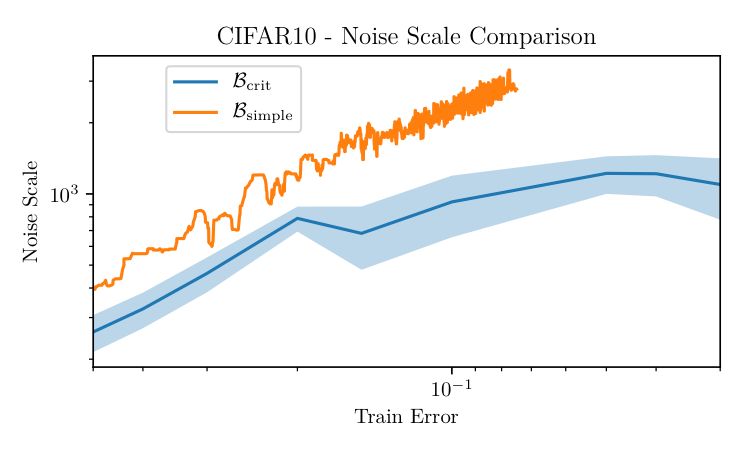}\\
 \includegraphics[width=0.5\textwidth]{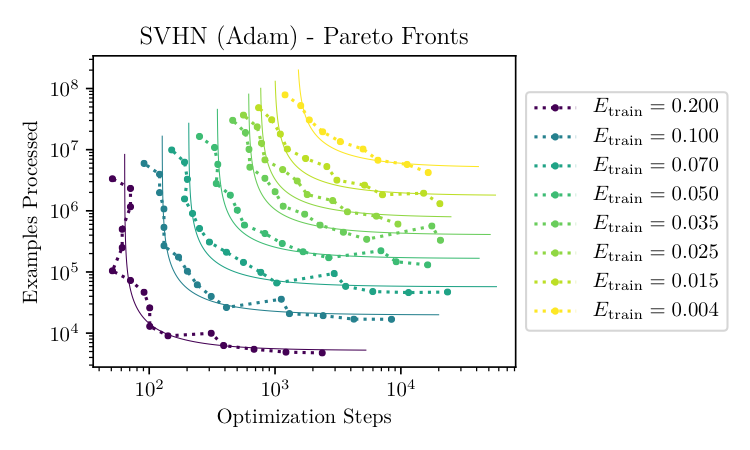}\includegraphics[width=0.5\textwidth]{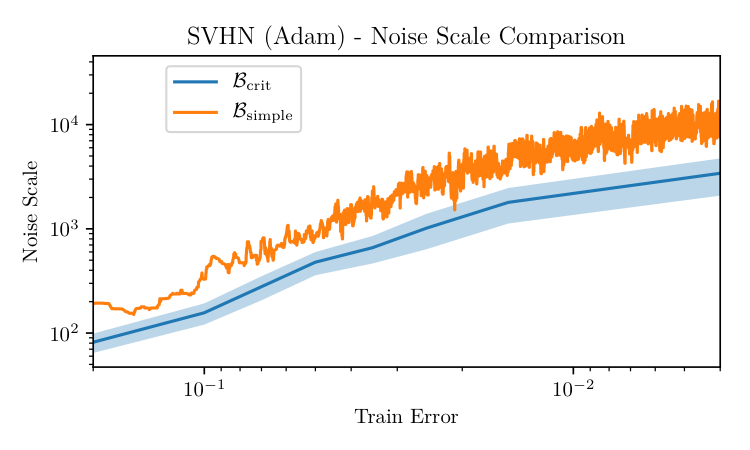}\\
 \includegraphics[width=0.5\textwidth]{\string"figures/Pareto_Fronts/Train/SVHN__SGD__-_Pareto_Fronts\string".png}\includegraphics[width=0.5\textwidth]{\string"figures/Noise_Comparisons/Train/SVHN__SGD__-_Noise_Scale_Comparison\string".png}\\
 \includegraphics[width=0.5\textwidth]{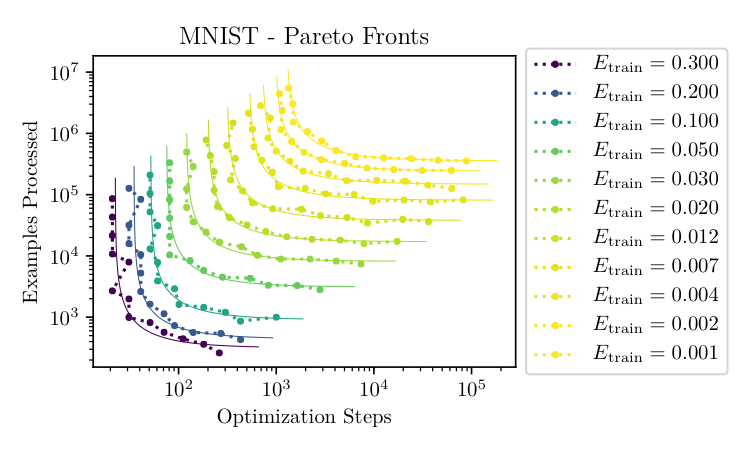}\includegraphics[width=0.5\textwidth]{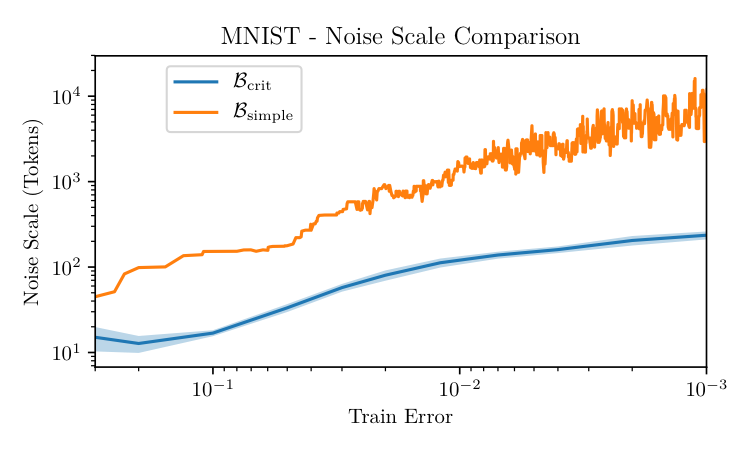}\caption{Scaling behavior of image classification tasks.\label{fig:classification-task-results}}
\end{figure}

In Figures \ref{fig:generative-modeling-task-results}, \ref{fig:imagenet-results}, \ref{fig:atari-task-results}, \ref{fig:atari-task-results-2}, \ref{fig:appdx-dota-results}, and \ref{fig:classification-task-results}, we display the results of a series of training runs for for classification, reinforcement learning, and generative modeling tasks. On the left, we show tradeoff curves between compute-efficiency and time-efficiency. Each point on each tradeoff curve represents the number of optimizer steps and processed training examples necessary to reach a given level of performance for a particular training run. Fits to the prediction of Equation \ref{eq:tradeoff-eqn} are shown. On the right, we compare the critical batch size, defined as the point where training is within 50\% of maximum efficiency in terms of both compute power and speed, and compare to the simple noise scale $\mathcal{B}_{{\rm simple}}$ of Equation \ref{eq:SimplestNoiseScale} and the true noise scale $\mathcal{B}_{{\rm noise}}$ of \ref{eq:noise-scale}, when available.  The results are summarized in Figure \ref{fig:noise-scale-summary} and table \ref{tab:noise-crit-comparison}.


\clearpage

\section{Temperature and the Noise Scale}
\label{subsec:temperature} \label{sec:Toy-Models-of}
\label{app:ToyModelsNS}


The noise scale measured during neural network training could depend on a variety of hyperparameters, such as the learning rate $\epsilon$ or momentum.  However, we have empirically found that noise scale primarily depends on $\epsilon$ and $B$ roughly through the ratio
\begin{eqnarray}
T(\epsilon, B) \equiv \frac{\epsilon}{\epsilon_{\rm max}(B)},
\end{eqnarray}
which we  refer to as the `temperature'  of training.   The terminology reflects an idea of the loss as a potential energy function, so that high temperature training explores a larger range of energies.

In the case of pure SGD it is approximated by $T \approx \epsilon/B$ in the small batch regime.  Our definition of $T$ can then be obtained from a toy model of a quadratic loss, which is described below.  In that case one can show explicitly  \cite{1704.04289} that the equilibrium distribution of gradients is characterized by this temperature\footnote{This definition can also be motivated by the empirical results of \cite{1711.00489}, which show that decaying the learning rate and increasing the batch size by the same factor have the same effect on training in the small-batch regime.}.

\begin{figure}
\includegraphics[width=0.5\textwidth]{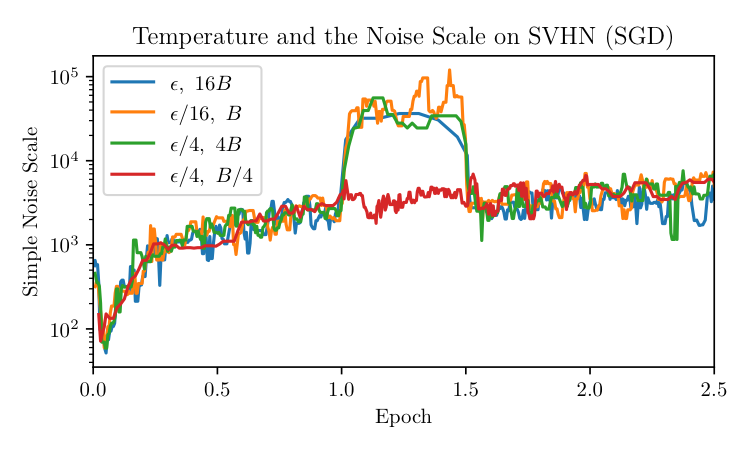}\includegraphics[width=0.5\textwidth]{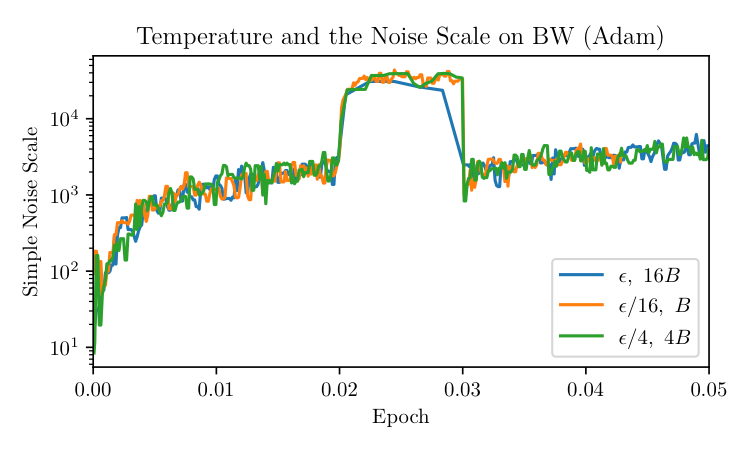}
 \caption{The noise scale is proportional to the inverse temperature.  On the left we display results for SVHN optimized via SGD, while on the right we have an LSTM on the Billion Word dataset optimized via Adam.  For each of the three curves, we modified either  the learning rate $\epsilon$, the batch size $B$, or both, so that the temperature $\frac{\epsilon}{B}$ was decreased by a factor of $16$ between epochs $1$ and $1.5$ (SVHN) or $0.02$ and $0.03$ (BW).  In all cases we see that the simple noise scale increased by a factor of $16$, then returned to roughly its original value once $ \epsilon, B$ were reset. } \label{fig:Temperature-Test}
\end{figure}

In equilibrium, the noise scales  vary in proportion to the inverse temperature, so that
\be
\CB_{\rm noise} \propto \CB_{\rm simple} \propto \frac{1}{T}.\label{eq:noise-vs-temperature}
\ee
It may seem surprising that higher temperature results in a smaller noise scale.  The intuition is that at larger $T$ the neural network parameters are further from the minimum of the loss, or higher up the `walls' of the potential, so that the gradient magnitude is larger relative to the variance.

Of course the loss landscape will be much more complicated than this toy model, but we have also observed that this scaling rule provides a good empirical rule of thumb, even away from pure SGD.  In particular, when we decay the learning rate $\epsilon$ by a constant factor, we often find that the noise scale grows by roughly the same factor.  ImageNet training provides an example in Figure \ref{fig:imagenet-results}.  A more direct investigation of the relation between $\CB_{\rm simple}$ and $T$ is provided in Figure \ref{fig:Temperature-Test}.

Since the noise depends primarily on the training temperature, and well-tuned training runs should have the same temperature at different batch sizes, the measured noise scale will also be consistent between optimally-tuned runs at different batch sizes.\footnote{This suggests that we can use the noise scale to \emph{define} the temperature via Equation \ref{eq:noise-vs-temperature}.  Then, once we have tuned the learning rate and measured the noise scale at small batch size, we can tune the learning rate at larger batch sizes to the noise scale constant.  Though we have not investigated this idea thoroughly, it could significantly simplify the problem of learning rate tuning at large batch size.}.  The noise scale then depends only on the temperature and the loss.

To summarize, the noise scale does \emph{not} provide an optimal training temperature schedule, but it instead prescribes an optimal batch size at any \emph{given} temperature.

\subsection*{A Toy Model for the Temperature}

Now let us consider a simple  explanation for the behavior of the noise scale in response to changes in the learning rate $\epsilon$ and batch size $B$.  
We start by approximating the loss as locally quadratic:
\[
L\left(\theta\right)=\frac{1}{2}\theta^{T}H\theta+{\rm const.}
\]
where we set $\theta=0$ at the minimum without loss of generality. To compute the noise scale, we need a model for the gradient covariance matrix $\Sigma$. A simple model appearing in \cite{conf/icml/SchaulZL13} suggests treating the per-example loss $L_{i}$ as a shifted version of the true loss, $L_{i}\left(\theta\right)=L\left(\theta-c_{i}\right)$, where $c_{i}$ is a random variable with mean zero and covariance matrix $\Sigma_{c}$. The gradient covariance matrix is then given by $\Sigma=H\Sigma_{c}H$, which is independent of $\theta$. The average gradient itself is given by $G=H\theta$, with $\theta$ changing in response to $\epsilon$ or $B$. As shown in \cite{1704.04289} over sufficiently long times SGD\footnote{With momentum, the same statements hold with $\epsilon$ replaced by $\epsilon/\left(1-m\right)$.} will approximately sample $\theta$ from the distribution
\[
p_{{\rm SGD}}\left(\theta\right)\propto\exp\left[-\frac{1}{2}\theta^{T}M^{-1}\theta\right]
\]
where the matrix $M$ satisfies
\[
MH+HM=\frac{\epsilon}{B}\Sigma.
\]

From these results, we can estimate the noise scale:
\begin{align*}
\mathcal{B}_{{\rm simple}} & =\frac{{\rm tr}(\Sigma)}{|G|^{2}}\approx\frac{B}{\epsilon}\frac{{\rm tr}\left(\Sigma\right)}{{\rm tr}\left(H^{2}\Sigma\right)}\\
\mathcal{B}_{{\rm noise}} & =\frac{{\rm tr}(\Sigma) H}{G^{T}HG}\approx\frac{B}{\epsilon}\frac{{\rm tr}\left(H\Sigma\right)}{{\rm tr}\left(H^{3}\Sigma\right)}
\end{align*}
So, in this model, the noise scale is expected to increase as we decrease the learning rate or increase the batch size.   We also expect that scaling the learning rate and batch size together should leave the noise scale unchanged.

When $B \ll \CB_{\rm simple}$, the ratio $\frac{\epsilon}{B}$  plays the role of a ``temperature''.  Since our analysis was only based on a toy model optimized using pure SGD, one might not expect it to work very well in practice.  However, as shown in Figure \ref{fig:Temperature-Test}, we have found that it provides a quite accurate model of the dependence of the noise scale on $\epsilon$ and  $B$ during neural network training, even when using the Adam\footnote{Note that with $\beta_2 = 0.999$ the Adam variance accumulators would take of order $\sim 1000$ steps to fully react.  On the right in Figure \ref{fig:Temperature-Test} we changed $\epsilon$ and $B$ for $0.01$ epochs, corresponding to between  $100$ and $1500$ optimizer steps.} optimizer.  For these tests, on SVHN we used an initial $(\epsilon, B) = (0.18, 128)$ while for billion word results we used $(\epsilon, B) = (6 \times 10^{-4}, 128)$.

Note that this result relies on the assumption that the optimizer has approached an effective equilibrium. We expect the equilibration timescale to be larger in the directions of low curvature, so that this effect will be strongest when the gradient points mostly in the large-curvature directions of the Hessian.  It would be interesting to investigate the timescale for equilibration.

\section{Dynamically Varying the Batch Size}

\label{app:DynamicBS}

As one can see from Figure \ref{fig:noise-scale-summary} and Section \ref{sec:EmpiricalResults}, both the measured $\mathcal{B}_{{\rm noise}}$ and $\mathcal{B}_{{\rm simple}}$, as well as the empirical $\mathcal{B}_{{\rm crit}}$ fit to Equation \ref{eq:tradeoff-eqn} all increase by at least an order of magnitude during training. Thus its natural to ask if we should expect to improve efficiency by dynamically scaling the batch size $B$ in response. We will see that the predicted gains are relatively modest unless the $\CB_{crit}$ changes greatly during training, although preliminary empirical tests suggest the benefits may be larger than predicted.

\subsection{Theory}

Consider a single full-batch optimizer step, over which the loss increases by an amount $\delta L$. If we instead use a batch of size $B$, it will take $\delta S=1+\frac{\mathcal{B}}{B}$ optimizer steps and $\delta E=B\delta S$ training examples to make the same amount of progress, where $\mathcal{B}$ is the noise scale.
Over a full training run, the total number of steps and data examples processed can be written as
\begin{align}
S & =\int\left(1+\frac{\mathcal{B}(s)}{B(s)}\right)ds\label{eq:total-steps-and-examples-apdx}\\
E & =\int\left(\mathcal{B}(s)+B(s)\right)ds\nonumber
\end{align}
where we parameterize the training trajectory by the number $s$ of full-batch optimizer steps (we abbreviated $S_{{\rm min}}$ above to $s$ for notational simplicity).

The question is how to optimally distribute the training examples over the full training trajectory. At each point along the trajectory, we have the choice of trading examples for optimizer steps by increasing or decreasing the batch size. This ``exchange rate'' between examples and steps is
\begin{equation}
r=-\frac{\frac{d}{dB}\delta E}{\frac{d}{dB}\delta S}=\frac{B^{2}(s)}{\mathcal{B}(s)}.
\end{equation}

If the distribution of training examples (and hence the batch size schedule) is optimal, then transferring examples from one part of training to another should not save any optimization steps. This means that \emph{the exchange rate $r$ should be constant throughout training}. Thus the batch size should be varied in proportion with the square root of the noise scale:
\begin{equation}
B(s)=\sqrt{r\mathcal{B}(s)}.\label{eq:best-variable-batch-size}
\end{equation}

We can determine the resultant Pareto front parameterizing the tradeoff between training cost and time by inserting Equation \ref{eq:best-variable-batch-size} into Equation \ref{eq:total-steps-and-examples-apdx} and eliminating the exchange rate\footnote{The exchange rate $r$ is a free parameter. It can be chosen according to preference from the value of training time vs compute.
There is also the fairly natural choice $r=\frac{E_{{\rm min}}}{S_{{\rm min}}}$ at which we have $\frac{S_{{\rm tot}}}{S_{{\rm min}}}=\frac{E_{{\rm tot}}}{E_{{\rm min}}}=1+\sqrt{\gamma}$, so that cost-efficiency and time-efficiency are both within the same factor of optimal, corresponding to the turning point in Figure \ref{fig:adaptive-batch-improvement}.}
\begin{align}
\frac{S_{{\rm tot}}}{S_{{\rm min}}}-1 & =\gamma\left(\frac{E_{{\rm tot}}}{E_{{\rm min}}}-1\right)^{-1},
\end{align}
where we define $S_{{\rm min}}\equiv\int ds$ and $E_{{\rm min}}\equiv\int\mathcal{B}ds$ to be the minimum possible number of optimizer steps and training examples needed to reach the desired level of performance, obtained by inserting $B\gg\mathcal{B}$ and $B\ll\mathcal{B}$ respectively into \ref{eq:best-variable-batch-size}. We also define
\begin{eqnarray}
\gamma & \equiv & \frac{\left(\int\sqrt{\mathcal{B}}ds\right)^{2}}{S_{{\rm min}}E_{{\rm min}}},\label{eq:gamma}
\end{eqnarray}
which parameterizes the amount of variation of the noise scale over the course of training. When the noise scale is constant $\gamma=1$ and there is no benefit from using an adaptive batch size; more variation in $\mathcal{B}$ pushes $\gamma$ closer to 0, yielding a corresponding predicted improvement in the Pareto front.\footnote{To see explicitly the dependence of $\gamma$ on the variability of the noise scale, we can rewrite it as $\gamma=\frac{\mathbb{E}\left[\sqrt{\mathcal{B}}\right]^{2}}{\mathbb{E}\left[\mathcal{B}\right]}=\frac{1}{1+\sigma_{\sqrt{\mathcal{B}}}^{2}/\mathbb{E}\left[\sqrt{\mathcal{B}}\right]^{2}}$, where the expectation is over a training run, weighting each full-batch step equally.} Note that since $\gamma$ involves the variation in the square root of $\CB$, practically speaking $\CB$ must vary quite a bit during training for adaptive batch sizes to provide efficiency benefits via these effects. Adaptive batch sizes may also have other benefits, such as replacing adaptive learning rates \cite{1711.00489} and managing the proportion of gradient noise during training.

\subsection{An SVHN Case Study}

\begin{figure}
\noindent \centering{}
\includegraphics[height=0.32\textwidth]{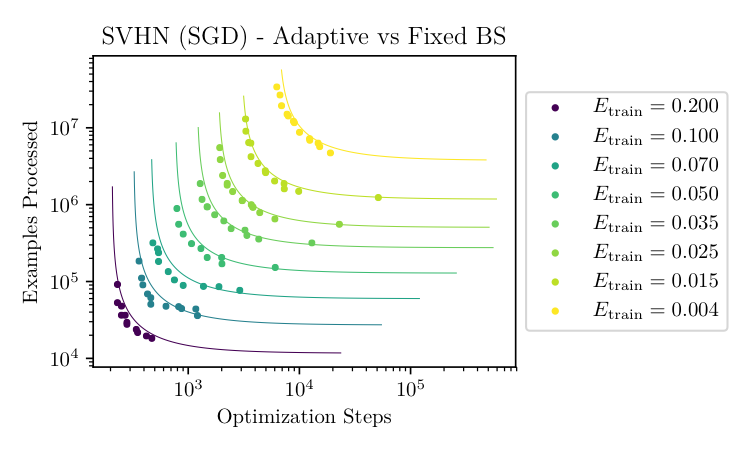}
\includegraphics[height=0.32\textwidth]{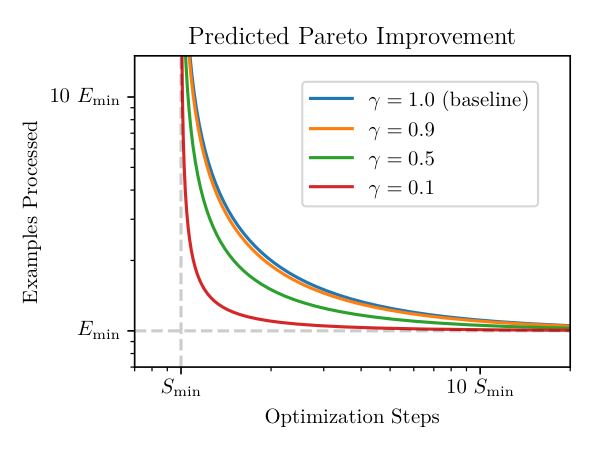}
\caption{\textbf{Left:} We compare training using an adaptive batch size (data points) to the hyperbolic fit to Pareto fronts at fixed batch size (lines). We note a modest but visible improvement to training efficiency. Adaptive batch sizes appear to decrease the minimum number of optimization steps $S_{{\rm min}}$, which was not anticipated by theoretical analysis. \textbf{Right:} Depending on the degree to which the noise scale varies over training, we can predict the potential Pareto improvement from using an adaptive batch size. \label{fig:adaptive-batch-improvement}}
\end{figure}
We have argued that a batch size of order the noise scale can simultaneously optimize data parallelism and total resource use. We have also shown that the noise scale tends to grow quite significantly during training. This suggests that one can further optimize resource use by adaptively scaling\footnote{This may have an additional advantage compared to training with a fixed, large batch size: it allows for a constant proportion of gradient noise during training, and some have argued \cite{1609.04836,1705.08741} that noise benefits generalization.} the batch size with the noise scale as training progresses, as discussed above.

For adaptive batch training, we can follow a simple and pragmatic procedure and dynamically set
\begin{eqnarray}
B=\sqrt{r\mathcal{B}_{{\rm simple}}},
\end{eqnarray}
with $\mathcal{B}_{{\rm simple}}$ measured periodically during training. 
The results from dynamic batch training with this procedure and various values of $r$ are compared to fixed batch size training in Figure \ref{fig:adaptive-batch-improvement}. We see that adaptive training produces a modest\footnote{It's challenging to provide a fair comparison between fixed and adaptive batch size training. Here we determined a roughly optimal relation $\epsilon=\frac{0.27B}{96+B}$ between learning rate $\epsilon$ and $B$ for fixed batch size training, and used this same function to determine the learning rate for both fixed and adaptive batch size training runs. This meant the adaptive batch size training used a corresponding adaptive learning rate. We did not experiment with learning rate schedules.} efficiency benefit.

We can combine our fixed batch size results with theoretical analysis to predict the magnitude of efficiency gains that we should expect from adaptive batch size training. We displayed $\mathcal{B}_{{\rm crit}}$ for fixed batch training of SVHN in Figure \ref{fig:svhn-pareto-front}. We have found that these results are fit very well by $\mathcal{B}_{{\rm crit}}(s)\approx10\sqrt{s}$, where $s$ is the number of steps taken in the limit of very large batch training. Using Equation \ref{eq:gamma}, we would predict the quite modest efficiency gain of
\begin{eqnarray}
\gamma=\frac{\left(\int ds\sqrt[4]{s}\right)^{2}}{s\int ds\sqrt{s}}=\frac{24}{25}
\end{eqnarray}
or around $4\%$. The benefits visible in Figure \ref{fig:adaptive-batch-improvement} in some cases appear too large to be fully explained by this analysis.

In particular, the adaptive batch size seems to benefit training in the regime of large batch size, decreasing the minimum number of optimization steps $S_{{\rm min}}$. However, our theoretical analysis would predict negligible benefits at large $E_{{\rm tot}}/S_{{\rm tot}}$. This may be due to the fact that the adaptive BS schedule also `warms up' the learning rate, or it may be an effect of a larger and more consistent proportion of gradient noise during training. It would be interesting to disentangle these and other factors in future work, and to study adaptive batch size training on other datasets.

\section{Comments on Optimization}

\begin{figure}
\noindent \centering{}\includegraphics[height=0.22\paperwidth]{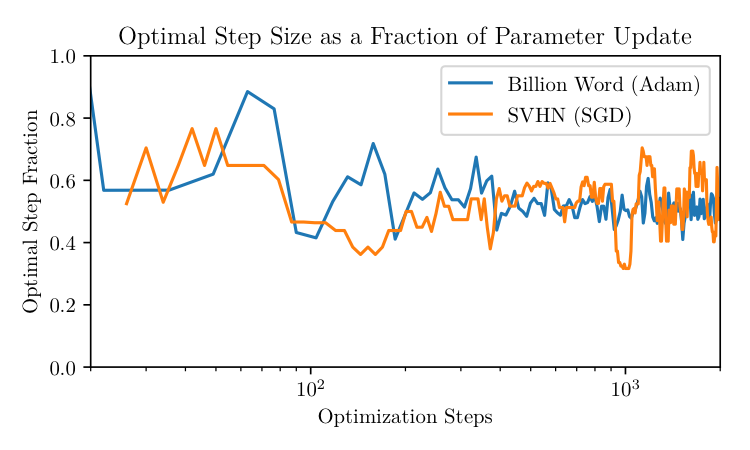}~~~\includegraphics[height=0.22\paperwidth]{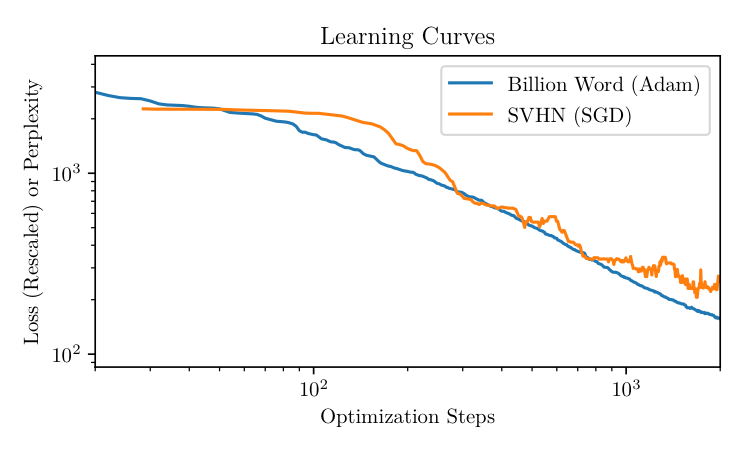} \caption{Left: This figure shows the magnitude of the optimal step size in the direction of the parameter update divided by the magnitude of the actual update. Optimal step sizes are determined by a line search of the loss. We show training of two quite different models with different optimizers -- an LSTM trained with Adam (momentum = $0.5$) on Billion Word, and a CNN trained on SVHN with SGD. In both cases, training converges to an approximate steady state where the average update is about twice the optimal update. Right: Learning curves included to clarify that this phenomenon is not due to the cessation of learning. \label{fig:optimal-step-fraction}}
\end{figure}
\begin{figure}
\noindent \centering{}\includegraphics[height=0.5\textwidth]{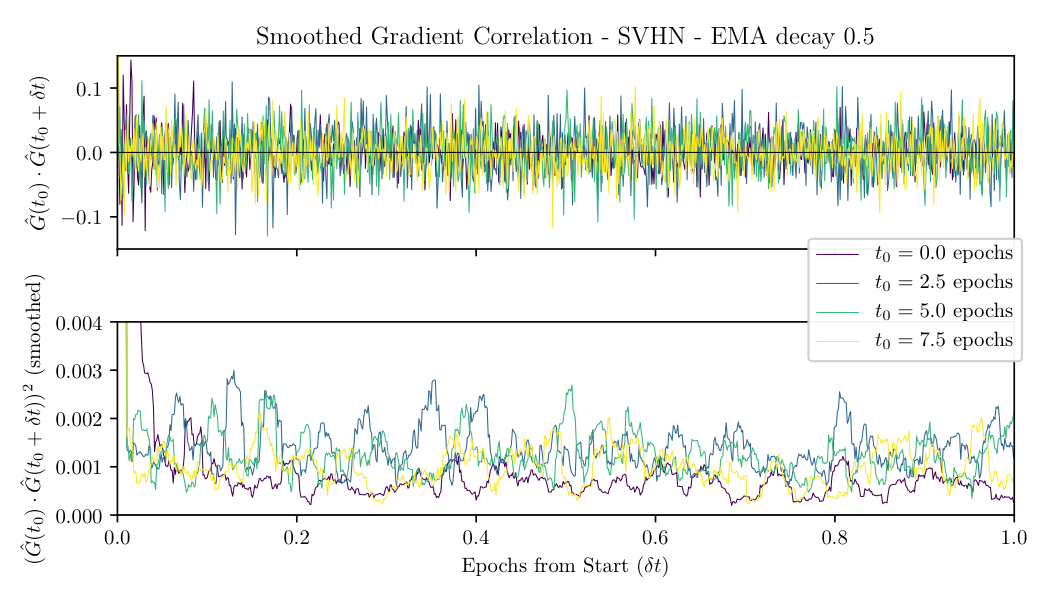}

\noindent \caption{The gradient exhibits rapid, long-lived oscillations over the course of training, even when using adaptive optimizers such as Adam. These oscillations are typical when optimizing functions with a large Hierarchy in the Hessian spectrum. We measure the moving average of the gradient with decay 0.5, computing its correlations over time. Results are shown for a simple CNN trained on SVHN using the Adam optimizer. \label{fig:gradient-correlation}}
\end{figure}

\subsection{Deterministic Training Performs Poorly}

\label{sec:BadDeterministicTraining}

From the discussion of the noise scale in Section \ref{sec:ParallelismandNoise}, we expect that at large batch size $B\gg\mathcal{B}_{\rm noise}$ we can obtain a very good estimate of the gradient. This would then suggest a minimally stochastic approach to training, where at each step one performs a line search of the true loss in the direction of the true gradient, and updates parameters accordingly.

This nearly deterministic `greedient descent' method performs poorly in practice \cite{1803.02021}.  While its first few steps tend to decrease the loss significantly, subsequent step sizes decrease rapidly and provide minimal further training progress.  In fact, when training with a fixed learning rate, we have observed that training often tends towards a regime where the optimal step size (determined by a line search in the direction of the parameter update) is almost exactly half of the actual update magnitude.  We have found that this phenomenon occurs regardless of the learning rate (scanning over several orders of magnitude), and seems common to a variety of different models, as shown in Figures \ref{fig:optimal-step-fraction} and \ref{fig:gradient-correlation}.
A natural interpretation of these results is that large Hessian directions are dominating the update \cite{unpublished-grd}, so that training involves rapid oscillations, as seen in Figure \ref{fig:gradient-correlation} (see \cite{goh2017why} for an intuitive picture of these oscillations).  Because of this, line searches do not appear to be useful in determining the optimal step size for a full training run.

\subsection{Motivations for Learning Rate Scaling Rules}
\label{app:ScalingRuleMotivation}

The learning rate scaling rule from Appendix \ref{sec:learning-rate-scaling} can be motivated as follows.  Equation \ref{eq:general-lr-scaling} generalizes Equation \ref{eq:OptimalStepwithNoise}, which was derived for plain SGD. The SGD linear scaling rule ($\alpha=1$) means that the step size per data example stays fixed up to $B_{\ast}$; one might intuitively expect that this is necessary to avoid diminishing returns as $B$ increases. In the case of Adam\footnote{These considerations apply equally well to RMSProp.}, we can use noise scale considerations to motivate the generalization to $0.5<\alpha<1$. The Adam update to the parameter $\theta_{i}$ takes the form
\[
\delta\theta_{i}=\epsilon\frac{\mathbb{E}_{\beta_{1}}\left[G_{i}\right]}{\sqrt{\mathbb{E}_{\beta_{2}}\left[G_{i}^{2}\right]}+\epsilon_{{\rm Adam}}}
\]
where $\mathbb{E}_{\beta}$ refers to an exponentially-weighted moving average with decay parameter $\beta$, and $G$ refers to a gradient from a batch of size $B$. If we disregard $\beta_{1},\beta_{2}$, and $\epsilon_{{\rm Adam}}$, this is roughly equivalent to
\[
\delta\theta_{i}\approx\epsilon\frac{{\rm sign}\left(\mathbb{E}\left[G_{i}\right]\right)}{\sqrt{1+\frac{s_{i}}{\mathbb{E}\left[G_{i}\right]^{2}}}}.
\]
where $s_{i}$ is the variance of $G_{i}$ over timesteps. If the step-to-step noise in the gradient is primarily due to batch statistics, $s_{i}$ should scale inversely with $B$. Comparing with \ref{eq:OptimalStepwithNoise}, this implies a square-root scaling rule ($\alpha=0.5$) to maintain a constant learning rate per data example. However, since $\beta_{2}$ is often set to large values around 0.999, the second moment accumulator may not have time to adapt to quick changes in the gradient noise; this pushes $\alpha$ back towards $1.0$. This may explain the variation in $\alpha$ between different tasks.

\subsection{Preliminary Tests of Generalization}
\label{app:generalization}
\begin{figure}
\includegraphics[width=0.5\textwidth]{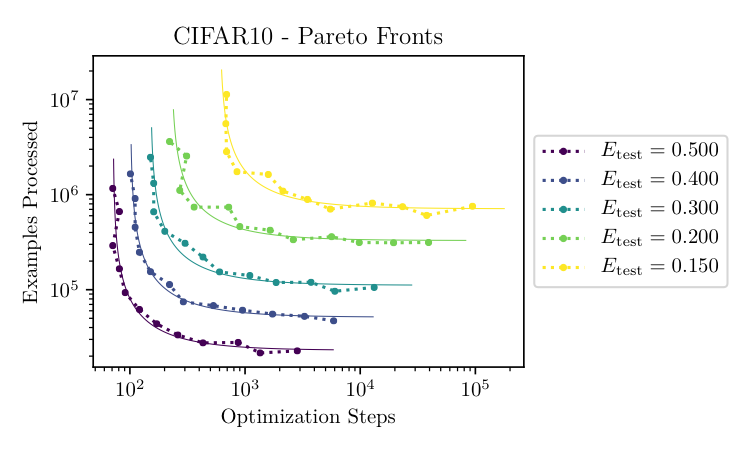}\includegraphics[width=0.5\textwidth]{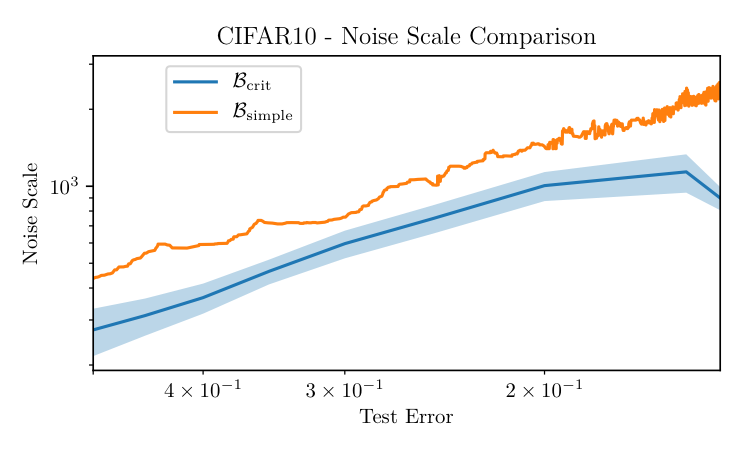}\\
 \includegraphics[width=0.5\textwidth]{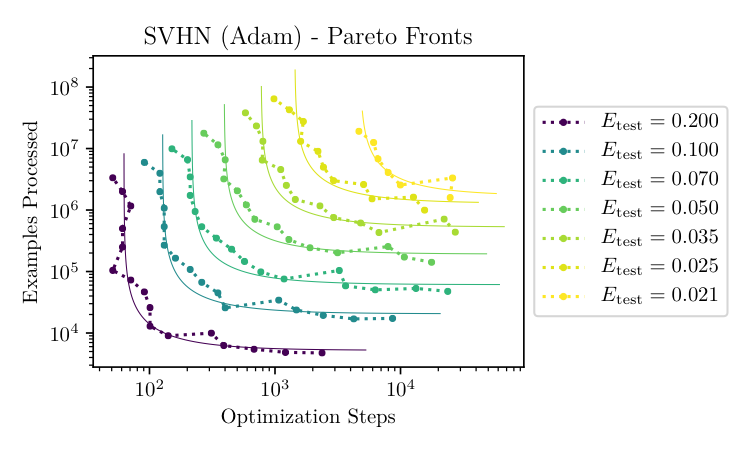}\includegraphics[width=0.5\textwidth]{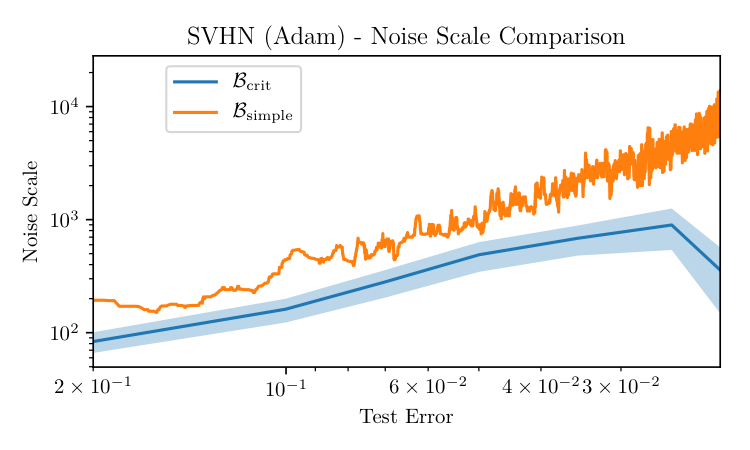}\\
 \includegraphics[width=0.5\textwidth]{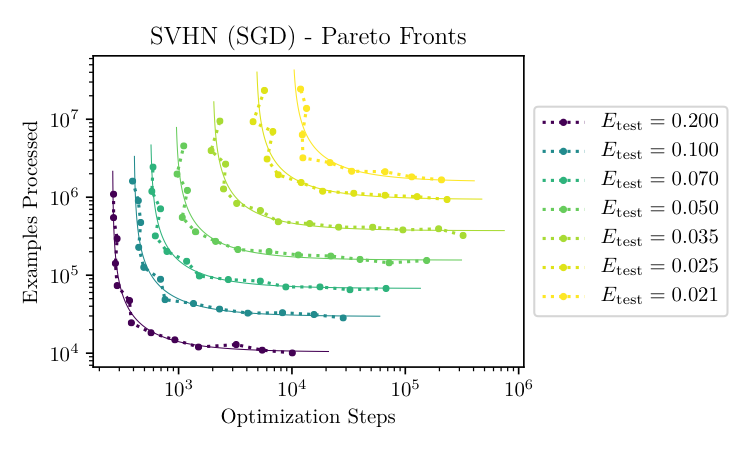}\includegraphics[width=0.5\textwidth]{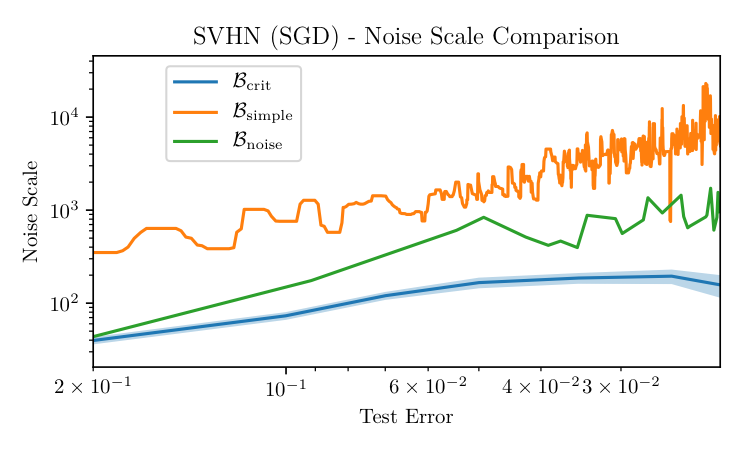}\\
 \includegraphics[width=0.5\textwidth]{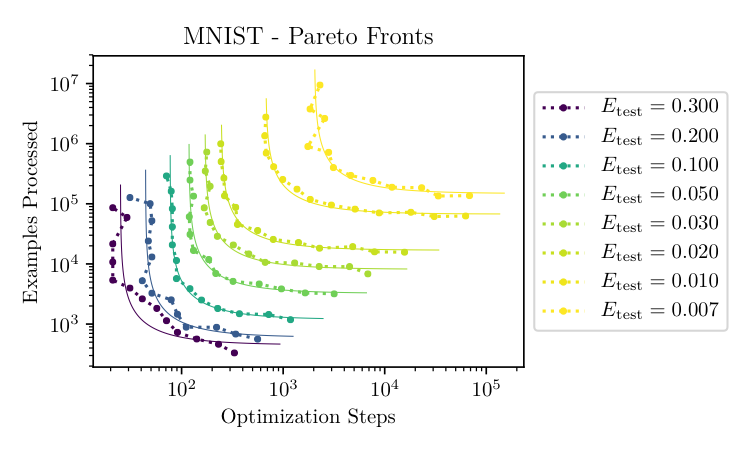}\includegraphics[width=0.5\textwidth]{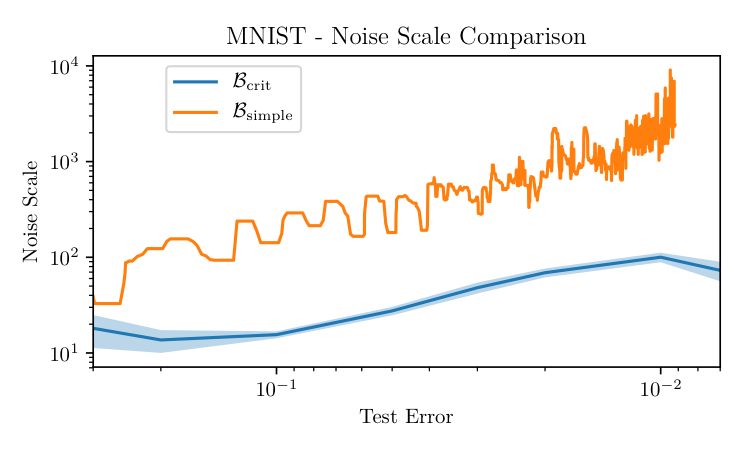}\caption{Scaling behavior of image classification tasks, using test set goals rather than Train set goals.  These results should be compared to those of Figure \ref{fig:classification-task-results}, which use training goals. \label{fig:classification-task-results-test}}
\end{figure}

Throughout the paper  we have  studied the noise scale as a function of the training loss or RL score.  But for non-RL tasks, it is also interesting to study the relationship between the noise scale and the critical batch size associated with minimization of the test loss.  The difference between the training and test results provides information about generalization.  In Figure \ref{fig:classification-task-results-test}  we report results using the test loss for the small image classification datasets.

As expected, early in training there is no difference between train and test results.  However, at the very end of training we observe a small  dip in $\CB_{\rm crit}$ for the test loss, which appears to occur consistently across datasets.  It would be very interesting to further investigate this phenomenon  in the future.

\pagebreak{}   \bibliographystyle{halpha}
\bibliography{bibliography}

\end{document}